\documentclass[11pt]{article}

\usepackage[utf8]{inputenc}
\usepackage[T1]{fontenc}
\usepackage[margin=1in]{geometry}
\usepackage[protrusion=true,expansion=false]{microtype}
\usepackage{amsmath,amssymb,bm}
\usepackage{booktabs}
\usepackage{multirow}
\usepackage{array}
\usepackage{placeins}
\usepackage{float}
\usepackage{graphicx}
\usepackage{caption}
\usepackage{subcaption}
\usepackage{xcolor}
\usepackage{pgfplots}
\usepgfplotslibrary{groupplots}
\usepackage{authblk}
\pgfplotsset{compat=1.18}
\usepackage[colorlinks=true,linkcolor=black,citecolor=black,urlcolor=blue]{hyperref}

\definecolor{fno}{RGB}{31,119,180}
\definecolor{don}{RGB}{127,127,127}
\definecolor{dpot}{RGB}{214,39,40}
\definecolor{gphyt}{RGB}{44,160,40}
\definecolor{poseidon}{RGB}{148,103,189}
\definecolor{walrus}{RGB}{255,127,14}
\definecolor{vicon}{RGB}{23,190,207}

\newcommand{\vrmse}{\mathrm{VRMSE}}

\newcommand{\rmae}{\mathrm{RMAE}}
\newcommand{\shortH}{\mathcal{H}_{\mathrm{s}}}
\newcommand{\fullH}{\mathcal{H}_{\mathrm{f}}}
\newcommand{\down}{\ensuremath{\downarrow}}
\newcommand{\figurepanelgraphic}[1]{%
  \IfFileExists{upload/#1}{%
    \includegraphics[width=\linewidth]{upload/#1}%
  }{%
    \IfFileExists{#1}{%
      \includegraphics[width=\linewidth]{#1}%
    }{%
      \fbox{\parbox[c][1.85in][c]{0.94\linewidth}{\centering
        {\scriptsize\ttfamily\detokenize{#1}}\\[3pt]
        Upload this PDF to Overleaf}}%
    }%
  }%
}
\title{\textbf{Beyond Local Accuracy:}\\[-1pt]
\textbf{Benchmarking the Rollout Reliability of PDE Foundation Models}}

\author[1]{Jaewan Park}
\author[1]{Seid Koric}
\author[2]{Giri Krishnan}
\author[2]{Sudhakar Pamidighantam}
\author[2]{Surresh Marru}
\affil[1]{National Center for Supercomputing Applications (NCSA)\\
University of Illinois Urbana--Champaign}
\affil[2]{Georgia Institute of Technology}
\date{August 2026}

\begin{document}
\maketitle

\begin{abstract}
PDE foundation models enable data-efficient adaptation but are deployed autoregressively, so short-horizon accuracy need not imply reliable simulation. We ask whether the short-horizon advantage of a foundation-model recipe persists when its predictions are repeatedly fed back as inputs for a long horizon. We define error amplification as $G=E(\fullH)/E(\shortH)$, which measures full-horizon error relative to short-horizon error. We evaluate seven models on seven two-dimensional time-dependent PDE datasets: scratch-trained FNO and DeepONet; fine-tuned DPOT, GPhyT, Poseidon, and Walrus; and frozen in-context VICON. We report VRMSE, relative $L_2$, and relative MAE. Six datasets provide paired short- and full-horizon scores, while CE-RM contributes short-rollout sweeps only. At least one fine-tuned foundation model outperforms both scratch baselines in 26 of 30 short-rollout dataset--budget cells under every metric. Nevertheless, full-horizon VRMSE exceeds short-horizon VRMSE in 140 of 144 paired trained configurations, with a median error amplification of $2.04$. The count below VRMSE one falls from 104 to 53, and 12 full rollouts exceed ten although none does at the short horizon. Walrus achieves the lowest full-horizon VRMSE in 16 of 24 cells, yet its median error amplification is $2.70$. Relative $L_2$ and relative MAE show the same qualitative pattern. These results show that favorable short-horizon rankings can coexist with substantial error amplification. PDE foundation-model benchmarks should therefore report horizon-resolved absolute errors, amplification statistics, and failure incidence rather than a single local forecast score.
\end{abstract}

\section{Introduction}
\label{sec:introduction}

Learning solution operators for PDEs has shifted from task-specific neural operators toward models pretrained across equations, resolutions, and physical domains, a transition enabled in part by scalable transformer-based architectures. Poseidon, DPOT, GPhyT, VICON, and Walrus represent distinct realizations of this broader program, spanning supervised fine-tuning, autoregressive denoising pretraining, multiphysics sequence modeling, in-context operator inference, and large-scale cross-domain pretraining
\cite{herde2024poseidon,hao2024dpot,wiesner2025gphyt,cao2024vicon,mccabe2025walrus}.
Despite their differences, these models share a central promise: knowledge acquired during pretraining should reduce the data and adaptation effort required to model a new system.

In practice, this promise is usually evaluated by asking whether a pretrained model achieves similar-to-lower error than a model trained from scratch at the same downstream-data budget. Such a comparison establishes data-efficient adaptation only at the horizon over which the error is measured. To determine whether this advantage translates into a useful dynamical model, however, the adapted model must also be evaluated over the long autoregressive rollouts encountered in deployment. After the first prediction, a learned PDE surrogate is conditioned on states generated by the model itself rather than on ground-truth states drawn from the data distribution. Small phase, amplitude, or conservation errors therefore alter the next input, which in turn affects every subsequent prediction. A short-horizon advantage shows that pretraining improves local prediction at a given data budget; it does not establish that this benefit survives repeated self-conditioning over the full trajectory.

The distinction matters especially for foundation models. Pretraining may provide a useful local predictive prior while leaving the adapted dynamics poorly suited to repeated autoregressive prediction. Conversely, a model with mediocre short-horizon accuracy may appear stable simply because it predicts an overly smooth field near the data mean. We therefore define \emph{error amplification} as the ratio of full-horizon error to the corresponding short-horizon error. The short-horizon score measures local fidelity, whereas error amplification measures how strongly that error grows during rollout. Neither quantity is sufficient alone: a low short-horizon error may conceal subsequent divergence, while low amplification may merely reflect consistently poor predictions. A benchmark must report both to expose the transition from local accuracy to long-horizon behavior.

We therefore organize this study around one falsifiable question:
\begin{quote}
\emph{Does the short-horizon advantage of a PDE foundation model persist over a full autoregressive rollout?}
\end{quote}
We test this question on six datasets with paired short- and full-horizon evaluations: four simulated flow settings from RealPDEBench-fluid-structure interaction (FSI), cylinder flow, controlled cylinder flow, and foil flow~\cite{hu2026realpdebench}-together with Diffusion--Reaction (DR) from PDEBench~\cite{takamoto2022pdebench} and conditioned incompressible Navier-Stokes (CondNS) from PDEArena~\cite{gupta2022pdearena}. 
We additionally include CE-RM from PDEGym~\cite{herde2024poseidon} as a complementary data-scaling study. Although its 21-frame trajectories do not support the paired horizon analysis, CE-RM uniquely extends the training-and-adaptation sweep to $N=1024$, whereas the other datasets end at $N=32$ or $64$. It therefore tests whether short-horizon model rankings and foundation-model advantages observed under small trajectory budgets persist when substantially more downstream data are available. CE-RM is included in the short-rollout data-efficiency summaries but excluded from all
horizon-amplification statistics. The resulting suite spans two- to four-field systems, square and rectangular grids, interpolation and Reynolds-extrapolation splits, and multiple rollout lengths. Across all evaluations, each model retains its native temporal configuration: the scoring interval specifies only the frames over which error is averaged and does not alter the model's input history or prediction window.

The main result is not simply which model ranks first. Walrus achieves the lowest error in most model-dataset-budget cells, and the identity of the best-performing model is more often preserved than changed across horizons. Yet error amplification exceeds one in nearly every paired configuration, indicating that absolute fidelity deteriorates even when the ranking remains stable. Moreover, dozens of configurations that outperform the mean-field reference at the short horizon become worse than that reference during full rollout, crossing from VRMSE below one to VRMSE above one. Ranking stability is therefore not equivalent to rollout reliability. The relevant object of evaluation is not foundation-model accuracy as a single scalar, but an accuracy-horizon profile conditioned on the downstream dataset and adaptation recipe.

Our contributions are:
\begin{itemize}
\item A unified evaluation of seven PDE models on seven time-dependent PDE datasets, reporting 3 different aspects of errors: VRMSE, relative $L_2$, and RMAE.

\item A short-versus-full-horizon comparison benchmark on six datasets that showcases how full-trajectory errors are amplified from the short-trajectory rollout.

\item Giving an empirical evidence that physics pretraining (foundation-model) is substantial but conditional: at least one fine-tuned foundation model outperforms both scratch baselines in 87\% among all cases.

\item Planned public release of all 180 checkpoints trained or fine-tuned in our benchmark sweep.

\end{itemize}

\section{Related work}
\label{sec:related}

\paragraph{Neural operators for PDEs}
Neural operators seek to approximate mappings between function spaces, enabling a learned
surrogate to map input fields, initial conditions, coefficients, or forcing terms directly to
solution fields. DeepONet represents such a mapping by combining a branch network that encodes
the input function with a trunk network that encodes the output coordinates
~\cite{lu2021deeponet}. The Fourier Neural Operator (FNO) instead parameterizes global
field-to-field transformations using spectral convolutions in Fourier space
~\cite{li2021fno}. These architectures established two complementary approaches to
task-specific operator learning and remain widely used baselines for PDE surrogate modeling.

For time-dependent PDEs, a neural operator can be trained as a finite-window evolution map and
then repeatedly applied to generate a longer trajectory. This composition turns a local
predictor into an autoregressive simulator: predicted fields re-enter the model as future inputs,
so full-rollout behavior is not determined by single-call accuracy alone. In our benchmark,
DeepONet and FNO are trained from scratch on each downstream dataset and serve as two established
task-specific references for comparison with pretrained PDE foundation models.

\paragraph{PDE foundation models.}
PDE foundation models extend operator learning from task-specific training toward reusable
representations of physical dynamics. Rather than initializing a separate model from scratch for
every equation and dataset, they are pretrained on trajectories drawn from multiple equations,
parameter regimes, resolutions, or physical domains. The pretrained representation is then
transferred to a downstream system through weight fine-tuning or in-context conditioning. The
central promise is amortization: knowledge acquired during pretraining should reduce the
downstream data or adaptation effort required to model a new system. Multiple Physics
Pretraining (MPP) provided an early realization of this idea by training an autoregressive
transformer across heterogeneous physical systems using shared field embeddings and
normalization, followed by transfer to previously unseen systems~\cite{mccabe2023mpp}.

Subsequent models expanded this paradigm through different architectures and pretraining
objectives. DPOT combines a scalable Fourier-attention architecture with autoregressive
denoising pretraining and was trained at scales of up to 0.5 billion parameters using more than
100,000 trajectories from over ten PDE datasets~\cite{hao2024dpot}. Poseidon instead employs a
multiscale operator transformer with time-conditioned normalization and a training strategy
based on the semigroup structure of time-dependent PDEs; it is pretrained on fluid-dynamics
equations and transferred to a broader collection of downstream operators
~\cite{herde2024poseidon}. GPhyT further broadens the physical scope by training a general
transformer on 1.8~TB of simulation data spanning fluid--structure interaction, shock dynamics,
thermal convection, and multiphase systems. Its original formulation emphasizes inferring
governing dynamics from context, including systems not observed during training
~\cite{wiesner2025gphyt}.

VICON and Walrus represent two additional directions. VICON uses patch-wise vision-transformer
processing to infer operators from demonstrations supplied at inference time, allowing it to
adapt to different physical systems and temporal strides without updating its weights
~\cite{cao2024vicon}. Walrus pursues large-scale cross-domain pretraining for continuum dynamics,
covering nineteen scenarios across classical fluids, astrophysics, geoscience, rheology, plasma
physics, and acoustics, while introducing mechanisms intended to improve forecast stability and
support training across two- and three-dimensional data~\cite{mccabe2025walrus}. These models
therefore share a transfer objective but not a common adaptation interface. In our benchmark,
DPOT, GPhyT, Poseidon, and Walrus are fine-tuned across downstream trajectory budgets, whereas
VICON remains frozen and uses in-context demonstrations. We preserve this distinction and treat
VICON as an inference-only reference rather than as another point in the fine-tuning
data-efficiency sweep.

\paragraph{PDE datasets and benchmarks.}
Public PDE benchmarks have made it possible to compare learned surrogates across equations under
standardized data and evaluation protocols. PDEBench provides simulation data, generation code,
baseline implementations, and multiple error metrics for a broad collection of time-dependent
PDEs with varying initial conditions, boundary conditions, and equation parameters
~\cite{takamoto2022pdebench}. PDEArena focuses on fluid-dynamics problems and evaluates
generalization across physical parameters and spatiotemporal scales
~\cite{gupta2022pdearena}. APEBench further emphasizes autoregressive emulation by providing a
differentiable simulation framework covering 46 PDE configurations across one, two, and three
spatial dimensions, together with rollout-oriented evaluation protocols
~\cite{koehler2024apebench}.

More recent collections have increased both the scale and physical diversity of available
simulation data. The Well contains approximately 15~TB of simulations across 16 datasets,
spanning fluid dynamics, biological systems, acoustics, magnetohydrodynamics, and astrophysical
phenomena~\cite{ohana2024thewell}. The PDEGym collection released with Poseidon provides
pretraining data and 15 downstream operator-learning tasks covering incompressible and
compressible flows, wave and reaction--diffusion equations, and steady and time-dependent
problems~\cite{herde2024poseidon}. RealPDEBench addresses a different limitation by pairing
measurements from five real physical systems with corresponding numerical simulations, thereby
supporting explicit evaluation of simulation-to-reality transfer
~\cite{hu2026realpdebench}.

Our benchmark does not introduce another physical dataset. Instead, it combines seven
two-dimensional, time-dependent datasets from these existing resources under a common
cross-model evaluation. We use four simulated flow systems from RealPDEBench---FSI, cylinder,
controlled cylinder, and foil---together with Diffusion--Reaction from PDEBench, conditioned
incompressible Navier--Stokes from PDEArena, and compressible-Euler Richtmyer--Meshkov
(CE-RM) from PDEGym. The first six datasets contain sufficiently long trajectories for paired
short- and full-horizon evaluation. CE-RM instead provides a complementary trajectory-budget
sweep through $N=1024$, although its 21-frame trajectories preclude the paired full-horizon
analysis. The contribution is therefore a unified, horizon-resolved comparison across models,
metrics, and data budgets rather than the creation of new simulation data.

\paragraph{Conditional generality and horizon-resolved evaluation.}
Recent bias-aware evaluation shows that physics foundation models behave as conditional rather
than universal generalists: their performance varies with physical regime, temporal scale,
initial-condition complexity, model size, architecture, and distribution shift
~\cite{chu2026bias}. Complementary rollout-focused studies have shown that autoregressive PDE
surrogates can lose accuracy as their predictions are repeatedly fed back as inputs
~\cite{koehler2024apebench,lippe2023pderefiner}. Our study connects these two evaluation
perspectives. Within each paired comparison, we hold the model, downstream dataset, trajectory
budget, and trained checkpoint fixed and vary only the scoring horizon. This design quantifies
error amplification from the short to the full horizon and tests whether the short-horizon
advantage of foundation-model pretraining persists under repeated self-conditioning.

\section{Benchmark question and measurements}
\label{sec:measurement}

\subsection{Autoregressive rollout}

Let $\bm{x}_t\in\mathbb{R}^{H\times W\times C}$ denote a $C$-channel physical field at time
$t$.  The adapters retain the temporal history and prediction-window configuration prescribed by
each model's released implementation or reference paper.  If $\mathcal I_m(t)$ denotes the
model-specific history assembled for model $m$ at rollout time $t$, deployment has the form
\begin{equation}
    \widehat{\bm{x}}^{(m)}_{t+1:t+k_m}
    = f_{\theta_m}\!\left(\mathcal I_m(t)\right),
\end{equation}
where both the input-history length and the number $k_m$ of states emitted by one model call may
depend on the model.  Once prediction begins, generated states are returned to that adapter's
history rather than replaced by ground truth.
The first generated frame is evaluated on an input history drawn entirely from the dataset.
Later frames are evaluated on histories increasingly shaped by model error.  The experiment
therefore probes two related but distinct objects: the learned local transition and the
dynamical system produced by iterating it.

For the six datasets with horizon-resolved harvests, we define a short \emph{scoring interval}
$\shortH=\{10,\ldots,20\}$, containing 11 predicted frames.
The full window is $\fullH=\{10,\ldots,100\}$ for RealPDEBench and DR (91 frames), and
$\fullH=\{10,\ldots,55\}$ for CondNS (46 frames).  All comparisons of short and full error are
paired within a dataset and checkpoint; we do not interpret a 91-step score as directly
equivalent to a 46-step score.  These intervals specify where error is averaged; they do not
specify or standardize a model's input-history or per-call output window.  CE-RM trajectories
contain 21 frames and therefore contribute a short track only.  Its reported range follows each
model's native stride: FNO, DeepONet, and Poseidon use $\Delta t=2$, while DPOT, GPhyT, Walrus,
and VICON use $\Delta t=1$.

\subsection{Metrics}

For test trajectory $i$, predicted time $t$, and channel $c$, let spatial averaging over the
$H\times W$ grid be denoted by $\langle\cdot\rangle_{HW}$.  We compute
\begin{align}
  \vrmse_{itc}
    &= \sqrt{
       \frac{\left\langle
       (\widehat{\bm{x}}_{itc}-\bm{x}_{itc})^2
       \right\rangle_{HW}}
       {\operatorname{Var}_{HW}(\bm{x}_{itc})+\varepsilon}},\\
  \mathrm{rel}\text{-}L_{2,itc}
    &= \sqrt{
       \frac{\left\langle
       (\widehat{\bm{x}}_{itc}-\bm{x}_{itc})^2
       \right\rangle_{HW}}
       {\left\langle\bm{x}_{itc}^{2}\right\rangle_{HW}+\varepsilon}},\\
  \rmae_{itc}
    &= \frac{\left\langle
       |\widehat{\bm{x}}_{itc}-\bm{x}_{itc}|
       \right\rangle_{HW}}
       {\left\langle|\bm{x}_{itc}|\right\rangle_{HW}+\varepsilon},
\end{align}
with $\varepsilon=10^{-7}$ and sample variance (one degree-of-freedom correction) in
$\vrmse$.  For each trajectory, the metric is averaged over the selected times and physical
channels.  We report the median over test trajectories.  Every prediction is scored in raw
physical units after any model-specific de-normalization.

VRMSE is the primary metric because it compares error with the spatial variability of each
field.  A value near one means that the error is comparable to the field's own variation.
Relative $L_2$ and relative MAE serve as robustness checks against conclusions driven by the
VRMSE normalization. We use a metric value greater than ten only as an operational flag for
catastrophic rollout failure; it is not a formal stability certificate.

\subsection{Metric and Temporal Coverage}

For the six trained models---FNO, DeepONet, DPOT, GPhyT, Poseidon, and Walrus---all three
metrics are available at both scoring intervals for every
$(\text{dataset},N,\text{model})$ cell on FSI, cylinder, controlled cylinder, foil, DR, and
CondNS. VICON contributes one frozen in-context result per dataset at both scoring intervals
and is not indexed by a fine-tuning budget. The four additional Walrus full-pool endpoints
reported separately in Appendix~\ref{app:fullpool} also include all three metrics.

For CE-RM, all three metrics are available for the six trained models at every
$N\in{1,4,16,64,256,1024}$. VICON results are also reported at the corresponding nominal
$N$ labels, but they are frozen in-context evaluations rather than fine-tuned checkpoints and
are excluded from the trained-model $N$-sweep statistics. Because CE-RM trajectories contain
only 21 frames, the dataset contributes to short-rollout summaries but not to short-to-full
horizon-amplification statistics.

\subsection{Operational Meaning of Rollout Reliability}

We use \emph{rollout reliability} as a finite-horizon empirical concept rather than as a formal
stability guarantee. A reliable model configuration should not only achieve low error over
$\shortH$, but should also avoid severe degradation over $\fullH$. For any of the three metrics,
let $E(\mathcal H)$ denote the error aggregated over scoring interval $\mathcal H$. We define the
horizon-amplification factor as
\begin{equation}
G = \frac{E(\fullH)}{E(\shortH)}.
\end{equation}
Thus, $G>1$ indicates error amplification, $G\approx1$ indicates little change, and $G<1$
indicates a reduction in the aggregated error over the longer interval.

The amplification factor measures relative degradation, not absolute predictive quality, and
therefore cannot be interpreted alone. A model that is similarly inaccurate at both horizons can
have $G\approx1$, whereas an initially accurate model can have $G>1$ while retaining acceptable
full-horizon error. We therefore evaluate rollout behavior using the short- and full-horizon
errors together with amplification factors, winner changes, crossings of the error-one threshold,
and the incidence of operationally defined catastrophic failures.

\section{Benchmark design}
\label{sec:design}

\subsection{Datasets and splits}

Table~\ref{tab:datasets} summarizes the seven datasets used in the benchmark.  The four
RealPDEBench scenarios contain simulated CFD fields.  Their test sets are deliberately
extrapolative: the five lowest- and five highest-Reynolds-number trajectories are reserved for
testing, validation trajectories are selected from the interior of the Reynolds-number range,
and the remaining trajectories form the training pool.  FSI has 36 available training
trajectories and therefore ends its sweep at $N=32$; cylinder, controlled cylinder, and foil end
at $N=64$.

DR uses 100 held-out trajectory groups selected from the upper end of the available index range,
whereas CondNS uses its native 130-trajectory test split.  CE-RM uses a fixed set of 130 held-out
test trajectories and has a substantially larger training pool, allowing its sweep to extend to
$N=1024$.  CE-RM therefore provides the only comparison in a higher-data regime beyond
$N=64$.  Channel-wise normalization statistics are computed using training data only, without
access to validation or test trajectories.

Within each model-specific adaptation track, the $N$-sweep uses nested training subsets:
increasing $N$ adds trajectories while leaving the validation and test sets unchanged.  Exact
training-subset identities are not standardized across model families, because Walrus and the
other trained models use different subset-construction rules.  Comparisons at the same $N$
therefore align the nominal number of downstream training trajectories, but not necessarily the
individual trajectories used for adaptation.

\begin{table}[t]
\centering
\caption{The seven benchmark datasets.  All datasets provide VRMSE, relative $L_2$, and relative
MAE.  The first six provide short- and full-horizon scores, whereas CE-RM provides the
short-horizon track only.  RealPDEBench entries use simulated fields; no experimental
measurements are evaluated.}
\label{tab:datasets}
\small
\resizebox{\textwidth}{!}{%
\begin{tabular}{llllcccl}
\toprule
Dataset & Family & Grid & Physical channels & Frames & Test traj. & $N$ sweep
& Horizon coverage \\
\midrule
FSI & fluid--structure interaction & $128^2$ & $(u,v,p)$ & 101 & 10
& ${1,4,16,32}$ & short, full \\
Cylinder & cylinder flow & $128\times256$ & $(u,v,p)$ & 101 & 10
& ${1,4,16,64}$ & short, full \\
Controlled cylinder & actuated cylinder flow & $128\times256$ & $(u,v,p)$ & 101 & 10
& ${1,4,16,64}$ & short, full \\
Foil & flow past an airfoil & $128\times256$ & $(u,v,p)$ & 101 & 10
& ${1,4,16,64}$ & short, full \\
DR & reaction--diffusion & $128^2$ & activator, inhibitor & 101 & 100
& ${1,4,16,64}$ & short, full \\
CondNS & conditioned incompressible NS & $128^2$ & scalar, $v_x$, $v_y$ & 56 & 130
& ${1,4,16,64}$ & short, full \\
CE-RM & compressible Euler, Richtmyer--Meshkov & $128^2$ & $(\rho,u,v,p)$ & 21 & 130
& ${1,4,16,64,256,1024}$ & short only \\
\bottomrule
\end{tabular}}
\end{table}

\subsection{Models and adaptation tracks}

The seven evaluated models are listed in Table~\ref{tab:models}.  Six models produce a distinct
checkpoint at every $N$: FNO and DeepONet are trained from scratch, whereas DPOT, GPhyT,
Poseidon, and Walrus are initialized from released pretrained checkpoints and fine-tuned.
These runs use matched nominal trajectory budgets, common test trajectories, and identical
reported metric definitions, while retaining each model's native temporal input and prediction
windows.  Parameter count, optimization compute, and subset construction are not controlled
across models: Walrus uses deterministic nested subsets obtained after a seed-0 shuffle, whereas
the other trained models use the first $N$ trajectories in pool order.  We therefore compare
model-specific end-to-end adaptation recipes at matched nominal $N$, rather than interpreting
their differences as a causal effect of pretraining alone.

VICON constitutes a separate inference-only reference.  Its weights remain frozen, and it
performs in-context prediction using nine demonstration pairs and one query pair.  Because VICON
does not produce an $N$-specific checkpoint, we report one reference result per dataset and
exclude it from all six-model training-sweep statistics.

\begin{table}[t]
\centering
\caption{Models and adaptation regimes.  Parameter counts correspond to the configurations
evaluated in this benchmark and are approximate.  The six swept models produce one trained
checkpoint at every available trajectory budget; VICON is a frozen in-context reference.}
\label{tab:models}
\small
\resizebox{\textwidth}{!}{%
\begin{tabular}{llrllc}
\toprule
Model & Class & Parameters & Published pretraining source & Adaptation here & Track \\
\midrule
FNO & neural operator & 101M & none & from scratch at each $N$ & swept \\
DeepONet & neural operator & 68--70M & none & from scratch at each $N$ & swept \\
DPOT-L & pretrained operator & 509M & PDEBench, PDEArena, CFDBench
    & fine-tuned at each $N$ & swept \\
GPhyT-L & physics transformer & 385M & The Well and in-house simulations
    & fine-tuned at each $N$ & swept \\
Poseidon-L & pretrained operator transformer & 629M & PDEGym
    & fine-tuned at each $N$ & swept \\
Walrus & continuum foundation model & 1.3B & 19 continuum scenarios
    & fine-tuned at each $N$ & swept \\
\midrule
VICON & in-context operator model & 88M & PDEArena and PDEBench
    & frozen, in-context & reference \\
\bottomrule
\end{tabular}}
\end{table}

\subsection{Evaluation protocol}

Each model retains the context length, prediction window, temporal stride, and rollout procedure
specified by its released implementation or reference configuration.  For the six
horizon-resolved datasets, predictions are aligned with the stored dataset frames beginning at
frame 10.  The short and full scoring intervals determine only which predictions are averaged;
they do not alter the model-specific rollout procedure.

FNO, DeepONet, DPOT, GPhyT, Poseidon, and Walrus are scored at the native dataset resolution.
VICON instead operates on a fixed $128\times128$ grid and is evaluated as a separate reference
track.  For the rectangular cylinder, controlled-cylinder, and foil datasets, both the saved
VICON predictions and their corresponding ground truth have resolution $128\times128$ and are
scored directly on that grid; predictions are not upsampled to the native $128\times256$
resolution.  All predictions are scored in raw physical units, with FNO and DeepONet outputs
de-normalized before evaluation.  VICON's seven-channel outputs are restricted to the active
physical channels in the corresponding ground truth.  This gives zero-based channel indices
$[1,2,3]$ for the four RealPDEBench datasets and the dataset-specific physical channels for DR,
CondNS, and CE-RM.

For each model--dataset--budget run, we evaluate the checkpoint selected by its training
pipeline.  FNO, DeepONet, DPOT, and GPhyT select the checkpoint with the lowest one-step,
teacher-forced validation MSE.  Poseidon's Trainer selects the checkpoint with the lowest
one-step supervised validation loss.  Walrus follows its released fine-tuning configuration,
using $2{,}000$ sampled windows per epoch, and uniquely selects checkpoints using multi-step
validation-rollout loss with early stopping.  Checkpoint selection is therefore part of each
model's adaptation recipe rather than a standardized component of the benchmark.

DR and CondNS use their complete 64-trajectory training pools at $N=64$.  Four additional
Walrus checkpoints trained on complete RealPDEBench pools are reported separately in
Appendix~\ref{app:fullpool} and excluded from all matched-$N$ aggregate statistics and rankings.
Across the common sweep, the benchmark aligns dataset splits, nominal trajectory budgets, test
trajectories, and reported metrics while retaining each model's native temporal, optimization,
and checkpoint-selection procedure.

Frozen VICON uses nine intra-trajectory demonstration pairs and one query pair without updating
its weights or accessing an $N$-trajectory fine-tuning set.  Its reported result is therefore
independent of $N$, including on CE-RM.  Earlier CE-RM outputs obtained by fine-tuning VICON on
$N$ trajectories are not part of the frozen in-context reference and are excluded from the
benchmark.

\begin{figure}[t]
\centering
\begin{subfigure}[t]{0.48\textwidth}
  \centering
  \figurepanelgraphic{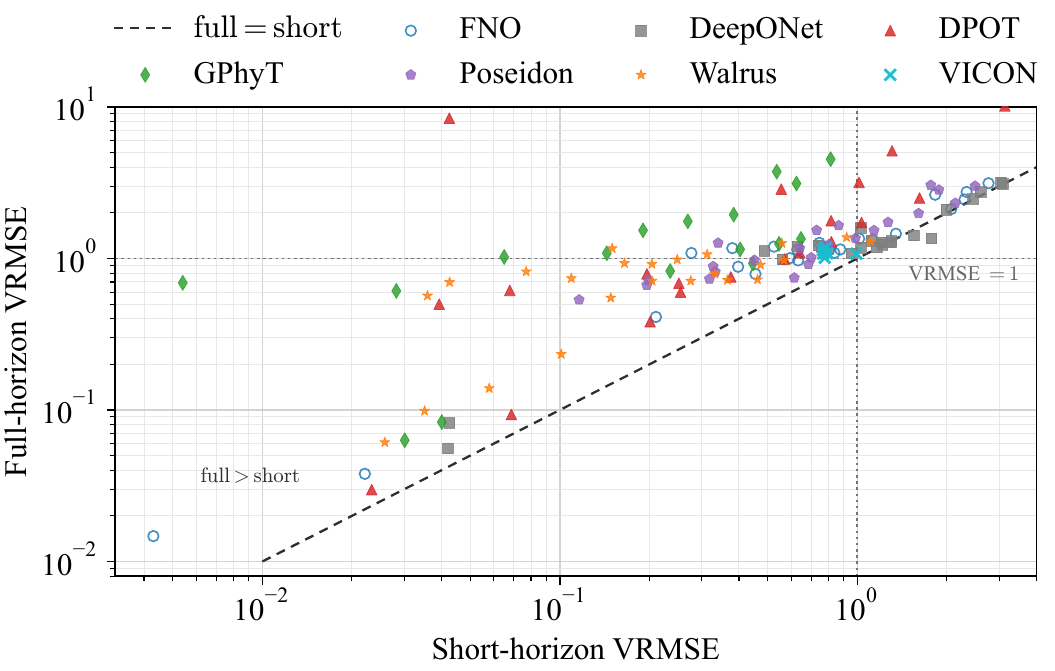}
  \caption{VRMSE.}
  \label{fig:shortfull-vrmse}
\end{subfigure}\hfill
\begin{subfigure}[t]{0.48\textwidth}
  \centering
  \figurepanelgraphic{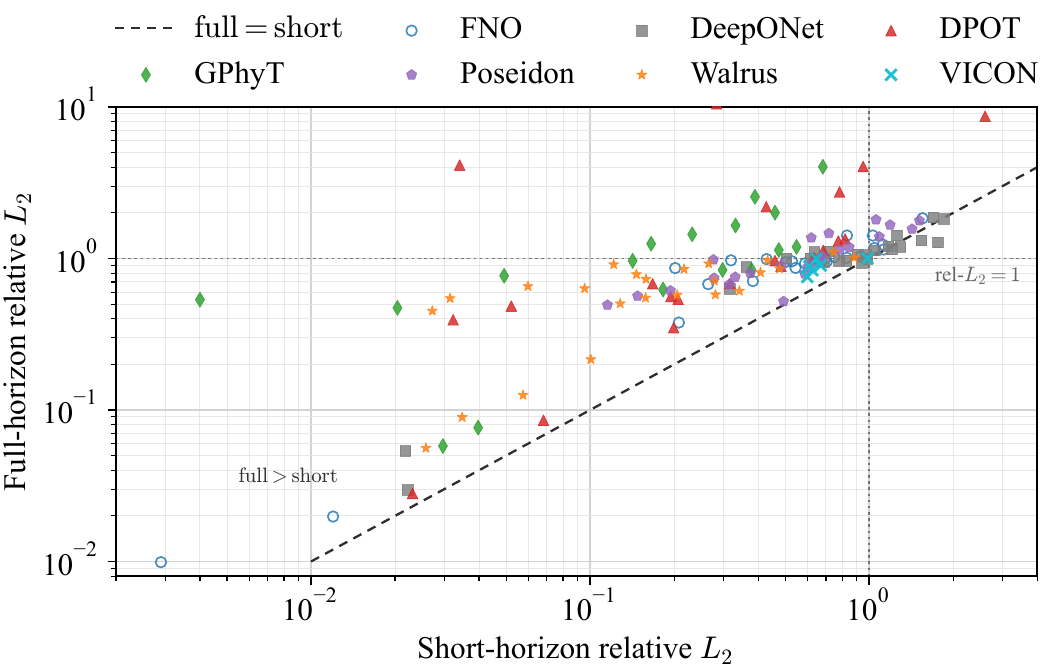}
  \caption{Relative $L_2$.}
  \label{fig:shortfull-rell2}
\end{subfigure}

\vspace{0.6em}
\begin{subfigure}[t]{0.48\textwidth}
  \centering
  \figurepanelgraphic{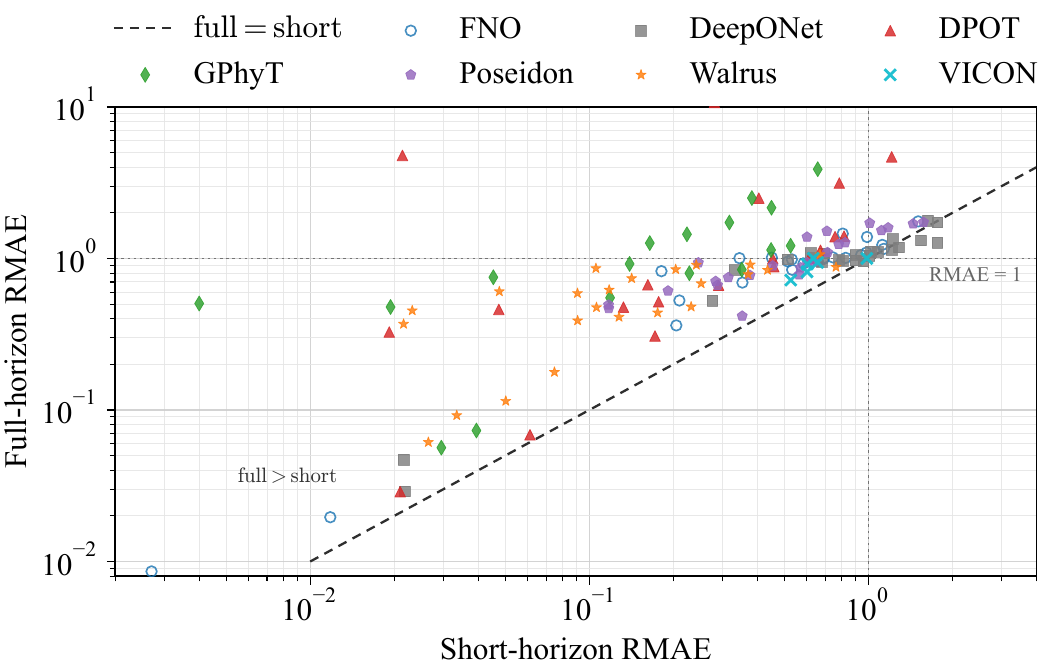}
  \caption{RMAE.}
  \label{fig:shortfull-rmae}
\end{subfigure}
\caption{Short- versus full-horizon errors on the six datasets with paired scoring intervals:
(a) VRMSE, (b) relative $L_2$, and (c) RMAE.  Each trained-model point pairs the same dataset,
budget $N$, and checkpoint; the six VICON points are frozen in-context references.  The dashed
diagonal marks equal error, and the dotted guides mark error one.  CE-RM is excluded because it
has no full-horizon score.  The full-horizon axis ends at $10^1$; respectively 12, 11, and 12
values in panels (a)--(c) are off-scale (diverged) and reported in the appendices.}
\label{fig:shortfull}
\end{figure}

\section{Results}
\label{sec:results}

\subsection{Longer horizons expose widespread error amplification}
\label{sec:horizon}

We first aggregate the complete swept track across datasets and trajectory budgets.  There are
30 short-rollout dataset--budget cells: 24 from the six horizon-resolved datasets and six from
CE-RM.  Each cell contains six trained models, yielding 180 short-rollout configurations per
metric.  The first 24 cells additionally provide full-horizon scores, yielding 144 paired
configurations.  Table~\ref{tab:aggregate} addresses four  questions:
\begin{enumerate}
    \item Does at least one fine-tuned FM outperform both from-scratch baselines?
    \item Is the short-horizon winner retained at the full horizon?
    \item How many metric values fall below value \textbf{one} (better than a constant field) or exceed \textbf{ten} (Too diverged)?
    \item How often does a pair of models reverse order between horizons?
\end{enumerate}
The first statistic compares the lower envelope of four fine-tuned models with that of two
scratch models.  It is an existence comparison, not an average-model comparison or a causal
estimate of pretraining, and the unequal group sizes should be kept in mind.

\begin{table}[t]
\centering
\caption{Aggregate diagnostics for the six swept models.  S/F denotes short/full scoring
intervals.  The short FM-envelope count includes all seven datasets; its full count uses the
six datasets with paired horizons.  All other horizon comparisons use 24 paired
dataset--budget cells and 144 trained-model configurations.  Pairwise reversals are measured
over $24\binom{6}{2}=360$ comparisons.}
\label{tab:aggregate}
\small
\resizebox{\textwidth}{!}{%
\begin{tabular}{lccccc}
\toprule
Metric &
FM envelope better (S/F), out of 30/24 &
Winner retained &
Paired $E<1$ (S/F), out of 144/144 &
Paired $E>10$ (S/F), out of 144/144 &
Pairwise reversals \\
\midrule
VRMSE & $26/21$ & $18/24$ & $104/53$ & $0/12$ & $104/360$ \\
Relative $L_2$ & $26/21$ & $16/24$ & $121/81$ & $0/11$ & $104/360$ \\
Relative MAE & $26/21$ & $19/24$ & $121/80$ & $0/12$ & $108/360$ \\
\bottomrule
\end{tabular}}
\end{table}

The lower-envelope result favors foundation-model recipes: at least one fine-tuned FM
outperforms both scratch baselines in 26 of 30 short-rollout cells and 21 of 24 full-horizon
cells under every metric.  Winner identity is also relatively persistent: the VRMSE winner is
retained in 18 of 24 cells, including 15 cells in which Walrus is best at both horizons.
Nevertheless, 104 of 360 pairwise VRMSE orderings reverse, and ranking retention does not imply
that the retained winner remains accurate enough for simulation.

The threshold counts expose this distinction.  Across all seven short tracks, 140 of 180 VRMSE
configurations fall below one.  Within the directly paired subset, however, this count decreases
from 104 of 144 at the short horizon to 53 of 144 at the full horizon.  Fifty-one configurations
cross from below one to at least one, whereas none crosses in the favorable direction.
Full-horizon VRMSE exceeds short-horizon VRMSE in 140 of 144 configurations, with median
amplification $G=2.04$.  No short-horizon VRMSE exceeds ten, whereas 12 full-horizon values do.
Relative $L_2$ and relative MAE show the same qualitative pattern.  Thus, stable rankings can
coexist with substantial losses of absolute fidelity, and the horizon gap is not an artifact of
a single normalization.

Figure~\ref{fig:shortfull} shows this gap directly under all three benchmark metrics by pairing,
for each trained model, the short- and full-horizon errors of the same dataset, trajectory budget,
and checkpoint.
In every panel, a point on the dashed diagonal has unchanged error, whereas a point above it has
deteriorated during the longer autoregressive rollout.  The dotted unit guides provide a common
visual reference while retaining the metric-specific normalizations.  FNO, DeepONet, DPOT,
GPhyT, Poseidon, and Walrus each contribute their complete 24-cell $N$-sweep; VICON contributes
one frozen in-context reference per dataset and is shown for context rather than counted as an
$N$-sweep.  CE-RM is absent only because its 21-frame trajectories do not define the paired full
scoring interval.  The broad concentration above the diagonal is consistent across VRMSE,
relative $L_2$, and RMAE, showing that the horizon gap is not created by any single
normalization.  For legibility, the displayed full-horizon axes stop at $10^1$; 12 VRMSE, 11
relative-$L_2$, and 12 RMAE outcomes are above that visible range, with their exact magnitudes
retained in Appendices~\ref{app:fullvrmse}, \ref{app:rell2}, and \ref{app:rmae}.

\subsection{The horizon gap persists at the largest matched data budget}
\label{sec:maxdata}

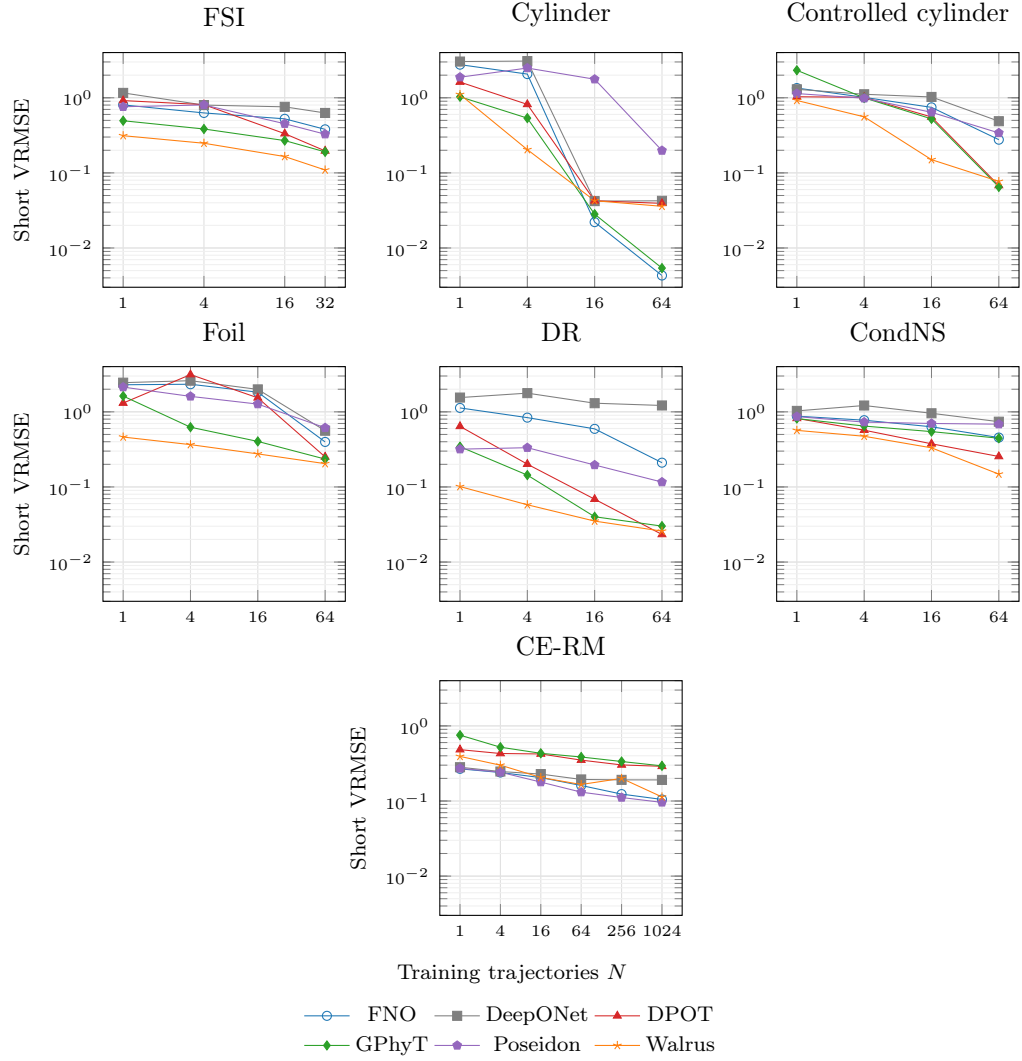
\begin{figure}[p]
\centering
\begin{tikzpicture}
\begin{groupplot}[
    group style={group size=3 by 3,horizontal sep=1.25cm,vertical sep=1.05cm},
    width=0.29\textwidth,
    height=0.205\textheight,
    xmode=log,
    log basis x=2,
    ymode=log,
    ymin=3e-3,
    ymax=4,
    title style={font=\small},
    label style={font=\scriptsize},
    tick label style={font=\tiny},
    grid=both,
    minor grid style={gray!12},
    major grid style={gray!25},
    every axis plot/.append style={mark size=1.7pt},
    legend columns=3,
    legend style={font=\scriptsize,draw=none}
]
\nextgroupplot[title={FSI},ylabel={Short VRMSE},xtick={1,4,16,32},xticklabels={1,4,16,32},legend to name=modellegend]
\addplot[color=fno,mark=o] coordinates {(1,0.8044) (4,0.6284) (16,0.5246) (32,0.3800)};
\addlegendentry{FNO}
\addplot[color=don,mark=square*] coordinates {(1,1.1615) (4,0.7997) (16,0.7576) (32,0.6279)};
\addlegendentry{DeepONet}
\addplot[color=dpot,mark=triangle*] coordinates {(1,0.9176) (4,0.8101) (16,0.3329) (32,0.1963)};
\addlegendentry{DPOT}
\addplot[color=gphyt,mark=diamond*] coordinates {(1,0.4946) (4,0.3841) (16,0.2693) (32,0.1904)};
\addlegendentry{GPhyT}
\addplot[color=poseidon,mark=pentagon*] coordinates {(1,0.7599) (4,0.8025) (16,0.4513) (32,0.3276)};
\addlegendentry{Poseidon}
\addplot[color=walrus,mark=star] coordinates {(1,0.3128) (4,0.2481) (16,0.1650) (32,0.1093)};
\addlegendentry{Walrus}

\nextgroupplot[title={Cylinder},xtick={1,4,16,64},xticklabels={1,4,16,64}]
\addplot[color=fno,mark=o,forget plot] coordinates {(1,2.7614) (4,2.0683) (16,0.0221) (64,0.0043)};
\addplot[color=don,mark=square*,forget plot] coordinates {(1,3.0381) (4,3.0814) (16,0.0421) (64,0.0425)};
\addplot[color=dpot,mark=triangle*,forget plot] coordinates {(1,1.6192) (4,0.8167) (16,0.0425) (64,0.0393)};
\addplot[color=gphyt,mark=diamond*,forget plot] coordinates {(1,1.0302) (4,0.5360) (16,0.0282) (64,0.0054)};
\addplot[color=poseidon,mark=pentagon*,forget plot] coordinates {(1,1.8800) (4,2.4884) (16,1.7664) (64,0.1984)};
\addplot[color=walrus,mark=star,forget plot] coordinates {(1,1.1067) (4,0.2040) (16,0.0426) (64,0.0359)};

\nextgroupplot[title={Controlled cylinder},xtick={1,4,16,64},xticklabels={1,4,16,64}]
\addplot[color=fno,mark=o,forget plot] coordinates {(1,1.3481) (4,1.0139) (16,0.7457) (64,0.2768)};
\addplot[color=don,mark=square*,forget plot] coordinates {(1,1.2966) (4,1.1138) (16,1.0255) (64,0.4875)};
\addplot[color=dpot,mark=triangle*,forget plot] coordinates {(1,1.0329) (4,1.0141) (16,0.5550) (64,0.0679)};
\addplot[color=gphyt,mark=diamond*,forget plot] coordinates {(1,2.3241) (4,0.9972) (16,0.5236) (64,0.0650)};
\addplot[color=poseidon,mark=pentagon*,forget plot] coordinates {(1,1.1381) (4,0.9858) (16,0.6406) (64,0.3405)};
\addplot[color=walrus,mark=star,forget plot] coordinates {(1,0.9209) (4,0.5576) (16,0.1500) (64,0.0771)};

\nextgroupplot[title={Foil},ylabel={Short VRMSE},xtick={1,4,16,64},xticklabels={1,4,16,64}]
\addplot[color=fno,mark=o,forget plot] coordinates {(1,2.2938) (4,2.3305) (16,1.8268) (64,0.3975)};
\addplot[color=don,mark=square*,forget plot] coordinates {(1,2.4458) (4,2.5983) (16,1.9907) (64,0.5605)};
\addplot[color=dpot,mark=triangle*,forget plot] coordinates {(1,1.3088) (4,3.1335) (16,1.5376) (64,0.2513)};
\addplot[color=gphyt,mark=diamond*,forget plot] coordinates {(1,1.6230) (4,0.6244) (16,0.4038) (64,0.2349)};
\addplot[color=poseidon,mark=pentagon*,forget plot] coordinates {(1,2.1347) (4,1.6069) (16,1.2683) (64,0.6138)};
\addplot[color=walrus,mark=star,forget plot] coordinates {(1,0.4618) (4,0.3660) (16,0.2755) (64,0.2042)};

\nextgroupplot[title={DR},xtick={1,4,16,64},xticklabels={1,4,16,64}]
\addplot[color=fno,mark=o,forget plot] coordinates {(1,1.1276) (4,0.8375) (16,0.5936) (64,0.2104)};
\addplot[color=don,mark=square*,forget plot] coordinates {(1,1.5492) (4,1.7741) (16,1.3031) (64,1.2155)};
\addplot[color=dpot,mark=triangle*,forget plot] coordinates {(1,0.6407) (4,0.2011) (16,0.0687) (64,0.0233)};
\addplot[color=gphyt,mark=diamond*,forget plot] coordinates {(1,0.3437) (4,0.1438) (16,0.0401) (64,0.0301)};
\addplot[color=poseidon,mark=pentagon*,forget plot] coordinates {(1,0.3184) (4,0.3331) (16,0.1960) (64,0.1161)};
\addplot[color=walrus,mark=star,forget plot] coordinates {(1,0.1011) (4,0.0579) (16,0.0351) (64,0.0258)};

\nextgroupplot[title={CondNS},xtick={1,4,16,64},xticklabels={1,4,16,64}]
\addplot[color=fno,mark=o,forget plot] coordinates {(1,0.8766) (4,0.7740) (16,0.6326) (64,0.4546)};
\addplot[color=don,mark=square*,forget plot] coordinates {(1,1.0292) (4,1.2140) (16,0.9568) (64,0.7396)};
\addplot[color=dpot,mark=triangle*,forget plot] coordinates {(1,0.8181) (4,0.5698) (16,0.3756) (64,0.2545)};
\addplot[color=gphyt,mark=diamond*,forget plot] coordinates {(1,0.8136) (4,0.6467) (16,0.5456) (64,0.4455)};
\addplot[color=poseidon,mark=pentagon*,forget plot] coordinates {(1,0.8649) (4,0.7298) (16,0.7003) (64,0.6850)};
\addplot[color=walrus,mark=star,forget plot] coordinates {(1,0.5667) (4,0.4729) (16,0.3318) (64,0.1483)};

\nextgroupplot[hide axis,xmin=1,xmax=2,ymin=1,ymax=2]
\nextgroupplot[title={CE-RM},ylabel={Short VRMSE},xtick={1,4,16,64,256,1024},
               xticklabels={1,4,16,64,256,1024}]
\addplot[color=fno,mark=o,forget plot] coordinates {(1,0.2681) (4,0.2396) (16,0.2050) (64,0.1609) (256,0.1236) (1024,0.1049)};
\addplot[color=don,mark=square*,forget plot] coordinates {(1,0.2817) (4,0.2469) (16,0.2283) (64,0.1937) (256,0.1912) (1024,0.1909)};
\addplot[color=dpot,mark=triangle*,forget plot] coordinates {(1,0.4828) (4,0.4290) (16,0.4229) (64,0.3503) (256,0.3022) (1024,0.2896)};
\addplot[color=gphyt,mark=diamond*,forget plot] coordinates {(1,0.7534) (4,0.5205) (16,0.4317) (64,0.3854) (256,0.3357) (1024,0.2937)};
\addplot[color=poseidon,mark=pentagon*,forget plot] coordinates {(1,0.2721) (4,0.2417) (16,0.1786) (64,0.1312) (256,0.1113) (1024,0.0959)};
\addplot[color=walrus,mark=star,forget plot] coordinates {(1,0.3925) (4,0.3001) (16,0.2057) (64,0.1659) (256,0.2001) (1024,0.1133)};
\end{groupplot}
\node[font=\scriptsize,anchor=north] at ([yshift=-0.12cm]current bounding box.south)
    {Training trajectories $N$};
\end{tikzpicture}

\ref{modellegend}
\caption{Short-horizon VRMSE versus the number of training trajectories for the six trained
models on all seven datasets.  Each model retains its native temporal configuration.  FSI
through CondNS are scored over frames 10--20; CE-RM uses the model-specific short rollout
defined in Section~\ref{sec:measurement}.  Frozen VICON is omitted because it is
$N$-independent.}
\label{fig:datasweep}
\end{figure}

Figure~\ref{fig:datasweep} shows short-horizon VRMSE across training budgets for all seven
datasets and six trained models.  VICON is omitted because its frozen in-context result is
independent of $N$.  Largest-matched-budget full-horizon results are summarized in
Table~\ref{tab:maxdata}, with the complete results reported in
Appendix~\ref{app:fullvrmse}.  Walrus is included at every available common $N$, including the
complete CE-RM sweep.

Table~\ref{tab:maxdata} reports results at the largest matched trajectory budget for each
dataset.  No single recipe wins every endpoint.  Walrus has the lowest full-horizon VRMSE on
FSI and CondNS, DPOT on controlled cylinder, foil, and DR, and scratch-trained FNO on cylinder;
Poseidon has the lowest CE-RM short-horizon VRMSE.  Among the six horizon-resolved datasets,
the short-horizon winner is retained on four: controlled cylinder changes from GPhyT to DPOT,
and foil changes from Walrus to DPOT.

The horizon gap nevertheless persists at these largest matched budgets.  Full-horizon VRMSE
exceeds short-horizon VRMSE in all 36 model--dataset configurations, with median amplification
$G=2.91$.  Even retained winners can accumulate substantial error: Walrus increases from
$0.1093$ to $0.7401$ on FSI and from $0.1483$ to $0.5498$ on CondNS.

Cylinder prevents an overly broad foundation-model conclusion.  At $N=64$, scratch-trained FNO
achieves short/full VRMSE of $0.0043/0.0147$.  Its short-horizon advantage over GPhyT
($0.0054$) is modest, but its full-horizon error is more than an order of magnitude lower than
that of every fine-tuned foundation-model recipe.  Thus, the benchmark supports a strong
cross-task foundation-model envelope without implying that a pretrained recipe must dominate
on every downstream system.

\begin{table}[t]
\centering
\caption{Results at the largest matched trajectory (largest $N$) budget for each dataset.  Entries are
short/full VRMSE for the six datasets with paired horizons and short VRMSE for CE-RM.  FSI
uses $N=32$, the next five datasets use $N=64$, and CE-RM uses $N=1024$.}
\label{tab:maxdata}
\small
\resizebox{\textwidth}{!}{%
\begin{tabular}{llrrrrrrr}
\toprule
Dataset & Report & $N$ & FNO & DeepONet & DPOT & GPhyT & Poseidon & Walrus \\
\midrule
FSI & VRMSE S/F & 32 & 0.3800/1.1699 & 0.6279/1.2005 & 0.1963/0.7915
    & 0.1904/1.5338 & 0.3276/0.8857 & 0.1093/0.7401 \\
Cylinder & VRMSE S/F & 64 & 0.0043/0.0147 & 0.0425/0.0819 & 0.0393/0.4999
    & 0.0054/0.6907 & 0.1984/0.7187 & 0.0359/0.5678 \\
Controlled cylinder & VRMSE S/F & 64 & 0.2768/1.0866 & 0.4875/1.1184 & 0.0679/0.6161
    & 0.0650/1.0230 & 0.3405/1.2643 & 0.0771/0.8203 \\
Foil & VRMSE S/F & 64 & 0.3975/0.8823 & 0.5605/0.9875 & 0.2513/0.6870
    & 0.2349/0.8261 & 0.6138/0.7450 & 0.2042/0.7104 \\
DR & VRMSE S/F & 64 & 0.2104/0.4113 & 1.2155/1.2340 & 0.0233/0.0298
    & 0.0301/0.0631 & 0.1161/0.5338 & 0.0258/0.0612 \\
CondNS & VRMSE S/F & 64 & 0.4546/0.7897 & 0.7396/1.2216 & 0.2545/0.6001
    & 0.4455/0.9285 & 0.6850/0.9155 & 0.1483/0.5498 \\
\addlinespace
CE-RM & VRMSE S & 1024 & 0.1049 & 0.1909 & 0.2896 & 0.2937 & 0.0959 & 0.1133 \\
\bottomrule
\end{tabular}}
\end{table}

\subsection{Data-efficiency trends and rollout failures are model-specific}
\label{sec:datasweep}

Figure~\ref{fig:datasweep} and Table~\ref{tab:shortvrmse} report short-horizon VRMSE across all
seven datasets and trajectory budgets.  Among the six trained models, Walrus has the lowest
error at $N=1$ on five of seven datasets and wins 19 of the 30 dataset--budget cells.  CE-RM is
an exception: FNO wins at $N=1$ and $N=4$, whereas Poseidon wins from $N=16$ through
$N=1024$.  Walrus improves with increasing $N$ on the other six datasets, but its CE-RM curve
is non-monotone, rising from $0.1659$ at $N=64$ to $0.2001$ at $N=256$ before decreasing to
$0.1133$ at $N=1024$.  Relative $L_2$ and RMAE exhibit the same CE-RM winner transition.

The short-horizon sweep alone does not expose model-specific rollout failures.  None of the 180
short-horizon configurations has VRMSE above ten, whereas 12 of the 144 paired full-horizon
configurations do.  The most extreme cases occur at $N=16$: DPOT reaches
$3.63\times10^6$ on foil, and GPhyT reaches $1.07\times10^{12}$ on controlled cylinder.
Consequently, the short-horizon curves and the complete full-horizon sweep in
Appendix~\ref{app:fullvrmse} answer different questions: the former measures local predictive
accuracy, whereas the latter tests whether that accuracy survives repeated self-conditioning.

Scratch-trained models can nevertheless dominate particular dataset--budget regimes.  FNO is
best on cylinder at $N=16$ and $N=64$ and on CE-RM at $N=1$ and $N=4$.  Model ordering is
therefore dataset-, budget-, and horizon-dependent rather than universal.

\begin{table}[p]
\centering
\caption{Short-horizon VRMSE ($\downarrow$) for all seven datasets.  FSI through CondNS are scored
over frames 10--20; CE-RM uses each model's native short rollout.  Frozen VICON contributes one
$N$-independent reference per dataset and is evaluated on its fixed $128\times128$ grid.}
\label{tab:shortvrmse}
\scriptsize
\resizebox{\textwidth}{!}{%
\begin{tabular}{llrrrrrrr}
\toprule
Dataset & $N$ & FNO & DeepONet & DPOT & GPhyT & Poseidon & Walrus & VICON$^{\dagger}$ \\
\midrule
\multirow{4}{*}{FSI} & 1 & 0.8044 & 1.1615 & 0.9176 & 0.4946 & 0.7599 & 0.3128 & \multirow{4}{*}{0.7677} \\
 & 4 & 0.6284 & 0.7997 & 0.8101 & 0.3841 & 0.8025 & 0.2481 & \\
 & 16 & 0.5246 & 0.7576 & 0.3329 & 0.2693 & 0.4513 & 0.1650 & \\
 & 32 & 0.3800 & 0.6279 & 0.1963 & 0.1904 & 0.3276 & 0.1093 & \\
\addlinespace
\multirow{4}{*}{Cylinder} & 1 & 2.7614 & 3.0381 & 1.6192 & 1.0302 & 1.8800 & 1.1067 & \multirow{4}{*}{0.7926} \\
 & 4 & 2.0683 & 3.0814 & 0.8167 & 0.5360 & 2.4884 & 0.2040 & \\
 & 16 & 0.0221 & 0.0421 & 0.0425 & 0.0282 & 1.7664 & 0.0426 & \\
 & 64 & 0.0043 & 0.0425 & 0.0393 & 0.0054 & 0.1984 & 0.0359 & \\
\addlinespace
\multirow{4}{*}{Controlled cylinder} & 1 & 1.3481 & 1.2966 & 1.0329 & 2.3241 & 1.1381 & 0.9209 & \multirow{4}{*}{0.7660} \\
 & 4 & 1.0139 & 1.1138 & 1.0141 & 0.9972 & 0.9858 & 0.5576 & \\
 & 16 & 0.7457 & 1.0255 & 0.5550 & 0.5236 & 0.6406 & 0.1500 & \\
 & 64 & 0.2768 & 0.4875 & 0.0679 & 0.0650 & 0.3405 & 0.0771 & \\
\addlinespace
\multirow{4}{*}{Foil} & 1 & 2.2938 & 2.4458 & 1.3088 & 1.6230 & 2.1347 & 0.4618 & \multirow{4}{*}{0.7773} \\
 & 4 & 2.3305 & 2.5983 & 3.1335 & 0.6244 & 1.6069 & 0.3660 & \\
 & 16 & 1.8268 & 1.9907 & 1.5376 & 0.4038 & 1.2683 & 0.2755 & \\
 & 64 & 0.3975 & 0.5605 & 0.2513 & 0.2349 & 0.6138 & 0.2042 & \\
\addlinespace
\multirow{4}{*}{DR} & 1 & 1.1276 & 1.5492 & 0.6407 & 0.3437 & 0.3184 & 0.1011 & \multirow{4}{*}{0.9933} \\
 & 4 & 0.8375 & 1.7741 & 0.2011 & 0.1438 & 0.3331 & 0.0579 & \\
 & 16 & 0.5936 & 1.3031 & 0.0687 & 0.0401 & 0.1960 & 0.0351 & \\
 & 64 & 0.2104 & 1.2155 & 0.0233 & 0.0301 & 0.1161 & 0.0258 & \\
\addlinespace
\multirow{4}{*}{CondNS} & 1 & 0.8766 & 1.0292 & 0.8181 & 0.8136 & 0.8649 & 0.5667 & \multirow{4}{*}{0.7832} \\
 & 4 & 0.7740 & 1.2140 & 0.5698 & 0.6467 & 0.7298 & 0.4729 & \\
 & 16 & 0.6326 & 0.9568 & 0.3756 & 0.5456 & 0.7003 & 0.3318 & \\
 & 64 & 0.4546 & 0.7396 & 0.2545 & 0.4455 & 0.6850 & 0.1483 & \\
\addlinespace
\multirow{6}{*}{CE-RM} & 1 & 0.2681 & 0.2817 & 0.4828 & 0.7534 & 0.2721 & 0.3925 & \multirow{6}{*}{0.7843} \\
 & 4 & 0.2396 & 0.2469 & 0.4290 & 0.5205 & 0.2417 & 0.3001 & \\
 & 16 & 0.2050 & 0.2283 & 0.4229 & 0.4317 & 0.1786 & 0.2057 & \\
 & 64 & 0.1609 & 0.1937 & 0.3503 & 0.3854 & 0.1312 & 0.1659 & \\
 & 256 & 0.1236 & 0.1912 & 0.3022 & 0.3357 & 0.1113 & 0.2001 & \\
 & 1024 & 0.1049 & 0.1909 & 0.2896 & 0.2937 & 0.0959 & 0.1133 & \\
\bottomrule
\end{tabular}}

\vspace{3pt}
\parbox{0.98\textwidth}{\footnotesize
$^{\dagger}$VICON~\cite{cao2024vicon} uses nine intra-trajectory demonstration pairs and one
query pair without weight updates.  It does not use the $N$-trajectory training pool, so one
$N$-independent value is reported per dataset.}
\end{table}

\subsection{Frozen in-context inference changes the question}
\label{sec:reference}

Table~\ref{tab:reference} reports VICON separately because it performs frozen in-context
inference without parameter updates.  On the six horizon-resolved datasets, its short-horizon
VRMSE ranges from $0.7660$ to $0.9933$, whereas all six full-horizon scores exceed one.  Its
median horizon amplification is comparatively modest at $G=1.38$, but this occurs because the
short-horizon error is already relatively high.  This illustrates why amplification cannot be
interpreted without absolute error: limited error growth does not necessarily imply accurate
simulation.

VICON therefore addresses a different question from the six trained models: whether
intra-trajectory demonstrations support prediction without task-specific weight updates, rather
than how performance scales with the number of fine-tuning trajectories.  Its result is
independent of $N$.  CE-RM contributes one frozen short-horizon reference value and no
full-horizon or $N$-sweep result.

\begin{table}[t]
\centering
\caption{Frozen in-context VICON VRMSE.  The first six datasets provide paired short- and
full-horizon scores; CE-RM provides a short-horizon score only.  VICON is evaluated on its
fixed $128\times128$ grid and does not define an $N$-sweep.}
\label{tab:reference}
\small
\begin{tabular}{lcc}
\toprule
Dataset & Short VRMSE & Full VRMSE \\
\midrule
FSI & 0.7677 & 1.0767 \\
Cylinder & 0.7926 & 1.0818 \\
Controlled cylinder & 0.7660 & 1.1361 \\
Foil & 0.7773 & 1.0114 \\
DR & 0.9933 & 1.0747 \\
CondNS & 0.7832 & 1.1758 \\
\addlinespace
CE-RM & 0.7843 & -- \\
\bottomrule
\end{tabular}
\end{table}

\section{Discussion}
\label{sec:discussion}

\subsection{What transfers and what does not}

The benchmark supports a limited but meaningful form of foundation-model transfer.  At least one
fine-tuned FM outperforms both scratch baselines in 26 of 30 short-horizon cells and 21 of 24
full-horizon cells under every metric.  The advantage therefore does not disappear when the
evaluation window is extended.  In practical terms, the available pretrained-and-adapted recipes
frequently provide a stronger downstream option than training FNO or DeepONet from scratch at the
same nominal trajectory budget.

This advantage is conditional rather than uniform.  Walrus accounts for most of the cellwise
wins, while other pretrained models vary substantially across datasets and budgets; scratch FNO
also remains the strongest model in particular regimes.  The evidence therefore does not support
a fixed hierarchy in which every pretrained model dominates every task-specific baseline.
Rather, the FM family expands the set of competitive downstream recipes, with the identity and
magnitude of the advantage depending on the target dynamics, data budget, and adaptation
procedure.

More importantly, relative advantage and rollout fidelity are distinct claims.  Foundation-model
recipes remain better than the scratch baselines in most full-horizon cells, yet full-horizon
VRMSE exceeds short-horizon VRMSE in 140 of 144 paired configurations.  These findings are not
contradictory: a model can lose substantial absolute fidelity while remaining the best available
option.  The results therefore support transfer of near-term predictive competence and, in many
cases, lower error over the full evaluation window relative to competing recipes.  They do not
show that successful downstream adaptation guarantees a faithful dynamical system under repeated
self-conditioning.


\subsection{What error amplification reveals}

Horizon amplification provides a paired measure of how a checkpoint changes when it is judged
over a longer rollout.  Because $E(\shortH)$ and $E(\fullH)$ are computed for the same model,
dataset, trajectory budget, and checkpoint, $G$ separates sensitivity to evaluation horizon from
differences between training runs.  This answers a question that a leaderboard cannot: not merely
which model has the lowest error, but how much of its near-term accuracy survives repeated
self-conditioning.

The resulting distinction is substantial.  Full-window VRMSE exceeds short-window VRMSE in 140
of 144 paired configurations, with a median amplification of $2.04$.  Among the 104 configurations
with short-horizon VRMSE below one, only 53 remain below one over the full window; the other 51
cross that reference level, and none crosses in the favorable direction.  Moreover, 12
full-horizon outcomes exceed VRMSE ten even though no short-horizon outcome does.  Similar
behavior under relative $L_2$ and RMAE shows that this pattern is not specific to variance
normalization.  Thus, the principal effect of extending the horizon is not limited to occasional
changes in model ranking: it can change whether a checkpoint remains useful in absolute terms.

At the same time, $G$ is not a standalone measure of rollout quality.  A model that is already
inaccurate over $\shortH$ can have $G\approx1$, whereas a model with a very small short-horizon
error can have a large ratio while retaining an acceptable full-horizon error.  The ratio is also
computed from averages over nested scoring intervals; it is neither a per-step error-growth rate
nor evidence that error increases monotonically at every frame.  Because the full intervals
differ across datasets, amplification should likewise be interpreted as a within-dataset
diagnostic rather than a universal stability coefficient.

Error amplification is therefore most informative when read together with the two absolute
errors.  The triplet $\bigl(E(\shortH),E(\fullH),G\bigr)$ distinguishes models that remain accurate, models that deteriorate from a strong starting point,
and models whose apparently small amplification merely reflects poor performance at both
horizons.  It provides a compact summary of horizon sensitivity, but not a formal certificate of
dynamical stability.

\subsection{Implications for model selection and benchmark design}

Checkpoint selection should be treated as part of an autoregressive model's evaluation protocol,
not as an implementation detail.  Five of the six swept recipes select checkpoints using
one-step supervised validation loss, whereas Walrus uses multi-step rollout validation and early
stopping.  The former therefore select models under a target that differs from their eventual
deployment as iterative simulators.  Our experiments do not establish that one-step selection
causes the observed horizon gap: Walrus also exhibits a median VRMSE amplification of $2.70$.
Nevertheless, the mismatch means that benchmark conclusions can reflect both the learned model
and the criterion used to select its checkpoint.

This distinction suggests two different standards.  A benchmark of released systems should
preserve each model's documented selection procedure, as done here, because that procedure is
part of the practical recipe.  By contrast, a controlled comparison of architectures or
pretraining strategies should standardize checkpoint selection and include validation over the
intended deployment horizon.  Simply selecting at the longest available horizon may be costly
and overly specific to that window; a multi-horizon validation profile is a more general design
target.  Whether such selection reduces amplification remains an empirical question rather than
a conclusion of the present benchmark.

The reported unit should likewise be an accuracy--horizon profile rather than a single forecast
score.  At minimum, the same checkpoint should be evaluated over paired near-term and deployment
windows, with both absolute errors and their amplification reported.  A fuller evaluation would
also provide error--time curves and per-trajectory distributions, which can reveal whether a
moderate aggregate error reflects gradual degradation or a small number of severe failures.
Model rankings alone are insufficient because a winner can be retained while every candidate
loses absolute fidelity.

Finally, a data-efficiency benchmark must specify more than the nominal number of trajectories.
Subset construction, optimization steps, parameter count, early stopping, and temporal adapters
can all differ at the same $N$.  Frozen in-context inference should also remain a separate track:
the number of demonstrations supplied at inference is not equivalent to the number of
trajectories used to update model weights.  Reporting these distinctions allows end-to-end
recipes to be compared transparently while preventing their differences from being misread as
the isolated effect of pretraining.

\subsection{Scope and limitations}

This benchmark compares complete model recipes rather than isolating the causal effect of
pretraining.  The models differ in architecture, parameter count, pretraining corpus,
optimization, subset construction, and checkpoint selection.  Matching the nominal trajectory
budget $N$ therefore equalizes the amount of downstream data only approximately; it does not
equalize the particular training samples, computational budget, or adaptation procedure.
Moreover, the FM-envelope statistic compares the best of four fine-tuned models with the best of
two scratch baselines.  It measures whether the FM family contains a strong downstream option,
not whether the average pretrained model is superior to the average scratch model.

The temporal comparisons also have a specific scope.  Short and full errors are paired within
each dataset and checkpoint, but the full interval contains 91 frames for RealPDEBench and DR
and 46 frames for CondNS.  Their amplification factors are therefore descriptive within-dataset
ratios, not universal error-growth rates that are directly comparable across physical systems.
CE-RM contributes only to the short-horizon analysis because its trajectories do not support
the longer scoring interval.  Retaining each model's native history, output window, and temporal
stride further makes this an evaluation of deployed recipes rather than a controlled ablation of
temporal architecture.

The empirical coverage is limited to seven two-dimensional, grid-based simulation datasets.
The RealPDEBench experiments use simulated fields only, so the results do not establish
robustness to experimental noise or sim-to-real transfer.  They also do not determine whether the
same horizon gap holds for three-dimensional systems, irregular discretizations, or substantially
longer deployment windows.  VICON should be interpreted only as a frozen in-context reference:
for the rectangular RealPDEBench cases it is evaluated on a $128\times128$ representation,
whereas the trained models are scored at the native $128\times256$ resolution.

Finally, the three reported metrics quantify field-level prediction error but do not identify its
physical or numerical cause.  They cannot by themselves distinguish phase drift, amplitude
error, conservation defects, or other failure mechanisms.  Likewise, the thresholds at errors
one and ten are operational diagnostics rather than mathematical stability criteria.  The
benchmark therefore establishes that conclusions about current PDE foundation-model recipes are
strongly horizon-dependent; it does not certify physical validity, prove formal instability, or
attribute the observed amplification to a particular modeling choice.

\section{Conclusion}
\label{sec:conclusion}

We evaluated seven PDE models across seven time-dependent systems using common definitions of
VRMSE, relative $L_2$, and RMAE while retaining each model's native temporal configuration.
The central question was whether the short-horizon advantage of foundation-model recipes
persists under autoregressive rollout.  The answer has two parts.  At least one fine-tuned FM
outperforms both scratch baselines in 26 of 30 short-horizon cells and 21 of 24 full-horizon
cells under every metric, showing that the relative downstream advantage often survives the
longer evaluation.  Absolute fidelity, however, deteriorates much more broadly: full-horizon
VRMSE exceeds short-horizon VRMSE in 140 of 144 paired configurations, with a median
amplification of $2.04$.  Fifty-one configurations cross from VRMSE below one to at least one,
and 12 exceed ten only over the full horizon.  The secondary metrics support the same qualitative
conclusion.

The key distinction is therefore between being better than competing recipes and remaining
accurate enough to function as a simulator.  Neither a low near-term error nor a retained
leaderboard rank establishes that repeated self-conditioning will remain faithful.  Horizon
amplification makes this loss visible, but it must be interpreted together with the corresponding
short- and full-horizon errors rather than as a standalone stability score.  PDE foundation
models should consequently be evaluated as the iterated dynamical systems that will be deployed,
using paired horizons, explicit failure reporting, and transparent adaptation and checkpoint-
selection protocols.  Because this study compares complete model recipes, its findings establish
a conditional empirical advantage for current foundation-model pipelines, not the isolated
causal effect of pretraining.

\section*{Reproducibility and supporting material}

The benchmark harness, model adapters, protocol documentation, and per-model instructions are
maintained at \url{https://github.com/cyber-shuttle/fm4pde}.  Each evaluated cell stores its
rollout archive and training log; the horizon-resolved harvest computes all three metrics from
saved predictions in raw units.  CE-RM contributes complete VRMSE, relative-$L_2$, and
relative-MAE sweeps for all seven models, including Walrus.  Runs used NCSA Delta and DeltaAI systems.  This work is supported by
the NSF Cybershuttle project (NSF Award No.\ 2209875).

\clearpage
\appendix

\section{Complete full-horizon VRMSE results}
\label{app:fullvrmse}

Table~\ref{tab:fullvrmse} contains the full-horizon counterpart to the common short-rollout sweep
in Table~\ref{tab:shortvrmse}.  It applies to the six datasets with 46- or 91-frame full intervals;
CE-RM has no full-horizon target because its trajectories contain 21 frames.

\begin{table}[!htbp]
\centering
\caption{Full-horizon VRMSE (\down) for the six datasets with a full interval.  RealPDEBench and
DR use frames 10--100; CondNS uses frames 10--55.}
\label{tab:fullvrmse}
\scriptsize
\resizebox{\textwidth}{!}{%
\begin{tabular}{lllrrrrrr}
\toprule
Dataset & Metric & $N$ & FNO & DeepONet & DPOT & GPhyT & Poseidon & Walrus \\
\midrule
FSI & Full VRMSE & 1 & 1.1338 & 1.1857 & 13.7525 & $5.27\times10^{3}$ & 1.1485 & 1.0644 \\
 & & 4 & 1.1458 & 1.1764 & 18.7319 & 1.9471 & 1.2374 & 0.9835 \\
 & & 16 & 1.1972 & 1.1950 & 16.6726 & 1.7582 & 0.9745 & 0.9301 \\
 & & 32 & 1.1699 & 1.2005 & 0.7915 & 1.5338 & 0.8857 & 0.7401 \\
\addlinespace
Cylinder & Full VRMSE & 1 & 3.1368 & 3.1663 & 2.5079 & $1.05\times10^{5}$ & 2.8369 & 1.2967 \\
 & & 4 & 2.1199 & 3.1016 & 1.7748 & 3.7422 & 3.0073 & 0.9177 \\
 & & 16 & 0.0379 & 0.0559 & 8.4288 & 0.6107 & 3.0491 & 0.6969 \\
 & & 64 & 0.0147 & 0.0819 & 0.4999 & 0.6907 & 0.7187 & 0.5678 \\
\addlinespace
Controlled cylinder & Full VRMSE & 1 & 1.4566 & 1.2790 & 1.7291 & $4.61\times10^{10}$ & 1.5327 & 1.3800 \\
 & & 4 & 1.3508 & 1.3107 & 3.1795 & $1.35\times10^{10}$ & 1.3579 & 1.2611 \\
 & & 16 & 1.2666 & 1.5879 & 2.8699 & $1.07\times10^{12}$ & 1.1667 & 1.1662 \\
 & & 64 & 1.0866 & 1.1184 & 0.6161 & 1.0230 & 1.2643 & 0.8203 \\
\addlinespace
Foil & Full VRMSE & 1 & 2.4533 & 2.4703 & 5.1371 & $3.34\times10^{7}$ & 2.3189 & 0.7262 \\
 & & 4 & 2.7470 & 2.7543 & 10.1646 & 3.1237 & 1.9837 & 0.7200 \\
 & & 16 & 2.6361 & 2.0992 & $3.63\times10^{6}$ & 1.1478 & 1.7347 & 0.7119 \\
 & & 64 & 0.8823 & 0.9875 & 0.6870 & 0.8261 & 0.7450 & 0.7104 \\
\addlinespace
DR & Full VRMSE & 1 & 1.3291 & 1.4241 & 1.0916 & 234.2 & 0.7326 & 0.2335 \\
 & & 4 & 1.0890 & 1.3610 & 0.3826 & 1.0827 & 0.8160 & 0.1391 \\
 & & 16 & 1.0046 & 1.3011 & 0.0934 & 0.0832 & 0.6663 & 0.0986 \\
 & & 64 & 0.4113 & 1.2340 & 0.0298 & 0.0631 & 0.5338 & 0.0612 \\
\addlinespace
CondNS & Full VRMSE & 1 & 1.1477 & 1.1776 & 1.2889 & 4.5224 & 1.6490 & 0.9750 \\
 & & 4 & 1.0720 & 1.2802 & 0.9927 & 1.3464 & 1.5362 & 0.9076 \\
 & & 16 & 0.9714 & 1.0832 & 0.7554 & 1.2490 & 1.0137 & 0.7919 \\
 & & 64 & 0.7897 & 1.2216 & 0.6001 & 0.9285 & 0.9155 & 0.5498 \\
\bottomrule
\end{tabular}}
\end{table}

\clearpage
\section{Complete relative-\texorpdfstring{$L_2$}{L2} results}
\label{app:rell2}

Table~\ref{tab:rell2all} reports relative $L_2$ for all seven datasets and every available
checkpoint.

\begin{table}[!htbp]
\centering
\caption{Relative $L_2$ (\down) for all seven datasets.  FSI through CondNS are short/full;
CE-RM entries are short only.}
\label{tab:rell2all}
\scriptsize
\resizebox{\textwidth}{!}{%
\begin{tabular}{llrrrrrr}
\toprule
Dataset & $N$ & FNO & DeepONet & DPOT & GPhyT & Poseidon & Walrus \\
\midrule
FSI & 1 & 0.7051/0.9567 & 0.9850/0.9984 & 0.7523/11.9418
    & 0.4321/$(4.5\times10^{3})$ & 0.6361/0.9414 & 0.2661/0.9251 \\
 & 4 & 0.5014/0.9468 & 0.6897/0.9515 & 0.6693/11.2274
    & 0.3320/1.6512 & 0.6787/1.0306 & 0.2171/0.8512 \\
 & 16 & 0.4291/0.9907 & 0.6172/0.9968 & 0.2840/10.5008
    & 0.2322/1.4436 & 0.3749/0.8001 & 0.1466/0.7855 \\
 & 32 & 0.3198/0.9724 & 0.5050/1.0007 & 0.1675/0.6819
    & 0.1655/1.2517 & 0.2781/0.7417 & 0.0956/0.6341 \\
\addlinespace
Cylinder & 1 & 1.5538/1.8431 & 1.6995/1.8670 & 0.8200/1.3470
    & 0.7195/$(6.76\times10^{4})$ & 1.1888/1.6645 & 0.8844/1.0123 \\
 & 4 & 1.1264/1.1462 & 1.8541/1.8243 & 0.4598/0.9685
    & 0.3897/2.5539 & 1.5138/1.7779 & 0.1586/0.7308 \\
 & 16 & 0.0120/0.0198 & 0.0222/0.0296 & 0.0341/4.1334
    & 0.0204/0.4721 & 1.0583/1.8022 & 0.0315/0.5462 \\
 & 64 & 0.0029/0.0099 & 0.0218/0.0536 & 0.0323/0.3937
    & 0.0040/0.5348 & 0.1480/0.5651 & 0.0272/0.4509 \\
\addlinespace
Controlled cylinder & 1 & 0.9302/1.0262 & 0.9438/0.9360 & 0.7756/1.3031
    & 1.7777/$(3.52\times10^{10})$ & 0.7837/1.1363 & 0.7469/1.1036 \\
 & 4 & 0.6935/0.9932 & 0.7795/0.9615 & 0.7830/2.7522
    & 0.7940/$(1.02\times10^{10})$ & 0.6604/0.9509 & 0.4326/0.9714 \\
 & 16 & 0.5292/0.9582 & 0.7183/1.1147 & 0.4277/2.1975
    & 0.4153/$(8.09\times10^{11})$ & 0.4843/0.8921 & 0.1215/0.9135 \\
 & 64 & 0.2018/0.8644 & 0.3633/0.8782 & 0.0522/0.4845
    & 0.0492/0.7693 & 0.2767/0.9813 & 0.0599/0.6552 \\
\addlinespace
Foil & 1 & 1.0442/1.1727 & 1.1723/1.2020 & 0.9515/4.0446
    & 1.0618/$(2.08\times10^{7})$ & 1.4244/1.5644 & 0.3428/0.6108 \\
 & 4 & 1.0302/1.4181 & 1.2594/1.4198 & 2.6035/8.6850
    & 0.4600/2.0066 & 1.0881/1.3907 & 0.2813/0.5775 \\
 & 16 & 0.8325/1.4185 & 0.9720/1.0555 & 1.2205/$(1.58\times10^{6})$
    & 0.2984/0.8393 & 0.8478/1.1880 & 0.2049/0.5745 \\
 & 64 & 0.2649/0.6769 & 0.3164/0.6273 & 0.1948/0.5615
    & 0.1826/0.6241 & 0.4935/0.5203 & 0.1581/0.5494 \\
\addlinespace
DR & 1 & 1.1194/1.2353 & 1.5386/1.3180 & 0.6342/1.0087
    & 0.3409/199.4 & 0.3154/0.6745 & 0.1006/0.2160 \\
 & 4 & 0.8311/1.0105 & 1.7654/1.2745 & 0.1994/0.3497
    & 0.1421/0.9649 & 0.3311/0.7537 & 0.0575/0.1254 \\
 & 16 & 0.5892/0.9292 & 1.2929/1.1921 & 0.0681/0.0856
    & 0.0397/0.0765 & 0.1940/0.6128 & 0.0348/0.0896 \\
 & 64 & 0.2080/0.3777 & 1.2076/1.1476 & 0.0231/0.0281
    & 0.0297/0.0578 & 0.1154/0.4925 & 0.0258/0.0561 \\
\addlinespace
CondNS & 1 & 0.7506/1.0205 & 0.9097/1.0645 & 0.6843/1.1321
    & 0.6825/4.0316 & 0.7172/1.4640 & 0.4780/0.8563 \\
 & 4 & 0.6739/0.9637 & 1.0482/1.1291 & 0.4832/0.8891
    & 0.5492/1.1929 & 0.6205/1.3693 & 0.4080/0.8086 \\
 & 16 & 0.5447/0.8665 & 0.8279/0.9686 & 0.3170/0.6832
    & 0.4745/1.1371 & 0.6030/0.8781 & 0.2787/0.7122 \\
 & 64 & 0.3831/0.7095 & 0.6360/1.1021 & 0.2068/0.5382
    & 0.3789/0.8402 & 0.5877/0.7926 & 0.1282/0.5038 \\
\addlinespace
CE-RM & 1 & 0.1887 & 0.1959 & 0.3241 & 0.4868 & 0.1918 & 0.2654 \\
 & 4 & 0.1686 & 0.1735 & 0.2945 & 0.3434 & 0.1709 & 0.2060 \\
 & 16 & 0.1438 & 0.1590 & 0.2915 & 0.2881 & 0.1254 & 0.1414 \\
 & 64 & 0.1128 & 0.1361 & 0.2409 & 0.2597 & 0.0921 & 0.1080 \\
 & 256 & 0.0858 & 0.1353 & 0.2115 & 0.2280 & 0.0782 & 0.1319 \\
 & 1024 & 0.0729 & 0.1346 & 0.2020 & 0.1979 & 0.0666 & 0.0761 \\
\bottomrule
\end{tabular}}
\end{table}

\clearpage
\section{Complete relative-MAE results}
\label{app:rmae}

Table~\ref{tab:rmaeall} reports relative MAE for every checkpoint.  The first six datasets provide
short/full pairs; CE-RM provides the complete short-rollout sweep.

\begin{table}[!htbp]
\centering
\caption{Relative MAE (\down) for all seven datasets.  FSI through CondNS are short/full;
CE-RM entries are short only.}
\label{tab:rmaeall}
\scriptsize
\resizebox{\textwidth}{!}{%
\begin{tabular}{llrrrrrr}
\toprule
Dataset & $N$ & FNO & DeepONet & DPOT & GPhyT & Poseidon & Walrus \\
\midrule
FSI & 1 & 0.6905/0.9912 & 1.0074/1.0425 & 0.8017/15.8779
    & 0.4137/$(2.14\times10^{3})$ & 0.6039/0.9370 & 0.2418/0.9069 \\
 & 4 & 0.5152/0.9550 & 0.6799/0.9479 & 0.7244/12.2328
    & 0.3176/1.7265 & 0.7137/1.0956 & 0.2040/0.8476 \\
 & 16 & 0.4509/1.0127 & 0.6295/1.0088 & 0.2807/10.6968
    & 0.2238/1.4445 & 0.3763/0.7745 & 0.1418/0.7395 \\
 & 32 & 0.3452/1.0051 & 0.5133/0.9863 & 0.1621/0.6702
    & 0.1647/1.2636 & 0.2836/0.7068 & 0.0907/0.5895 \\
\addlinespace
Cylinder & 1 & 1.5089/1.7577 & 1.6370/1.7790 & 0.8181/1.4050
    & 0.6862/$(3.58\times10^{4})$ & 1.1769/1.5955 & 0.7656/0.8778 \\
 & 4 & 1.1356/1.1580 & 1.7660/1.7353 & 0.4543/0.9977
    & 0.3824/2.5046 & 1.4466/1.6939 & 0.1178/0.6206 \\
 & 16 & 0.0118/0.0196 & 0.0218/0.0290 & 0.0214/4.7894
    & 0.0194/0.4785 & 1.0111/1.7124 & 0.0232/0.4533 \\
 & 64 & 0.0027/0.0086 & 0.0216/0.0470 & 0.0192/0.3271
    & 0.0040/0.5044 & 0.1174/0.4687 & 0.0216/0.3696 \\
\addlinespace
Controlled cylinder & 1 & 0.9414/1.0537 & 0.9591/0.9559 & 0.7596/1.3948
    & 1.7142/$(3.13\times10^{10})$ & 0.7829/1.2409 & 0.6709/1.0629 \\
 & 4 & 0.6927/1.0116 & 0.7840/0.9761 & 0.7868/3.1478
    & 0.6827/$(5.13\times10^{9})$ & 0.6388/0.9533 & 0.3775/0.9102 \\
 & 16 & 0.5318/0.9862 & 0.6960/1.0838 & 0.4050/2.5021
    & 0.3483/$(3.99\times10^{11})$ & 0.4542/0.8944 & 0.1056/0.8613 \\
 & 64 & 0.1813/0.8246 & 0.3317/0.8392 & 0.0474/0.4611
    & 0.0453/0.7515 & 0.2459/0.9378 & 0.0476/0.6056 \\
\addlinespace
Foil & 1 & 0.9846/1.0945 & 1.0839/1.1038 & 1.2121/4.6878
    & 1.1000/$(1.29\times10^{7})$ & 1.5760/1.7336 & 0.2317/0.4804 \\
 & 4 & 0.9887/1.3870 & 1.2213/1.3440 & 3.5365/13.0607
    & 0.4496/2.1620 & 1.1138/1.5296 & 0.1757/0.4388 \\
 & 16 & 0.8092/1.4596 & 0.9005/0.9700 & 1.3455/$(1.4\times10^{6})$
    & 0.2283/0.7991 & 0.8250/1.2726 & 0.1278/0.4111 \\
 & 64 & 0.2103/0.5281 & 0.2759/0.5267 & 0.1325/0.4776
    & 0.1187/0.5504 & 0.3530/0.4169 & 0.0908/0.3891 \\
\addlinespace
DR & 1 & 1.1217/1.2298 & 1.5448/1.3139 & 0.6134/0.9994
    & 0.3398/125.5 & 0.2880/0.6667 & 0.0750/0.1779 \\
 & 4 & 0.8292/1.0094 & 1.7663/1.2694 & 0.1719/0.3081
    & 0.1394/0.9207 & 0.3147/0.7483 & 0.0502/0.1146 \\
 & 16 & 0.5856/0.9269 & 1.2888/1.1837 & 0.0613/0.0686
    & 0.0394/0.0732 & 0.1914/0.6111 & 0.0335/0.0922 \\
 & 64 & 0.2050/0.3617 & 1.2111/1.1353 & 0.0210/0.0289
    & 0.0295/0.0564 & 0.1172/0.4937 & 0.0265/0.0613 \\
\addlinespace
CondNS & 1 & 0.7461/1.0167 & 0.8996/1.0540 & 0.6737/1.1330
    & 0.6584/3.8845 & 0.7101/1.5159 & 0.4351/0.8389 \\
 & 4 & 0.6590/0.9610 & 1.0212/1.1103 & 0.4583/0.8828
    & 0.5265/1.2134 & 0.6027/1.3856 & 0.3708/0.7856 \\
 & 16 & 0.5329/0.8427 & 0.8087/0.9604 & 0.2906/0.6664
    & 0.4481/1.1370 & 0.5718/0.8688 & 0.2516/0.6840 \\
 & 64 & 0.3537/0.6934 & 0.6212/1.0898 & 0.1771/0.5180
    & 0.3519/0.8439 & 0.5645/0.7802 & 0.1061/0.4755 \\
\addlinespace
CE-RM & 1 & 0.1219 & 0.1399 & 0.2562 & 0.4235 & 0.1278 & 0.2084 \\
 & 4 & 0.1106 & 0.1191 & 0.2280 & 0.2683 & 0.1151 & 0.1393 \\
 & 16 & 0.0943 & 0.1208 & 0.2037 & 0.2192 & 0.0830 & 0.0904 \\
 & 64 & 0.0748 & 0.0903 & 0.1685 & 0.1941 & 0.0582 & 0.0707 \\
 & 256 & 0.0578 & 0.0889 & 0.1496 & 0.1667 & 0.0492 & 0.0906 \\
 & 1024 & 0.0488 & 0.0890 & 0.1444 & 0.1441 & 0.0426 & 0.0467 \\
\bottomrule
\end{tabular}}
\end{table}

\clearpage
\section{Complete frozen in-context reference metrics}
\label{app:reference}

\begin{table}[!htbp]
\centering
\caption{VRMSE, relative $L_2$, and relative MAE for frozen in-context VICON across all seven
datasets.  The first six entries are short/full; CE-RM entries are short only.  This reference is
not included in the trained-model aggregate counts of Table~\ref{tab:aggregate}.}
\label{tab:referenceall}
\small
\begin{tabular}{lrrrr}
\toprule
Dataset & $N$ & VRMSE & rel-$L_2$ & RMAE \\
\midrule
FSI & -- & 0.7677/1.0767 & 0.6703/0.9002 & 0.6578/0.9411 \\
Cylinder & -- & 0.7926/1.0818 & 0.6259/0.8370 & 0.6030/0.8157 \\
Controlled cylinder & -- & 0.7660/1.1361 & 0.6097/0.8875 & 0.6042/0.9176 \\
Foil & -- & 0.7773/1.0114 & 0.6003/0.7558 & 0.5270/0.7202 \\
DR & -- & 0.9933/1.0747 & 0.9869/1.0010 & 0.9877/1.0009 \\
CondNS & -- & 0.7832/1.1758 & 0.6472/0.9962 & 0.6344/1.0032 \\
\addlinespace
CE-RM & 1 & 0.5384 & 0.3573 & 0.2798 \\
 & 4 & 0.4023 & 0.2721 & 0.1978 \\
 & 16 & 0.7802 & 0.5069 & 0.4729 \\
 & 64 & 0.3648 & 0.2478 & 0.1833 \\
 & 256 & 0.4157 & 0.2770 & 0.2156 \\
 & 1024 & 0.3343 & 0.2265 & 0.1681 \\
\bottomrule
\end{tabular}
\end{table}

\section{Additional Walrus full-pool endpoints}
\label{app:fullpool}

The common-$N$ Walrus checkpoints appear in every trained-model table and figure.  For four
RealPDEBench datasets, an additional checkpoint was fine-tuned on the complete pool beyond the
largest common sweep value.  Table~\ref{tab:fullpool} records those extra endpoints without using
them in any same-$N$ aggregate or ranking.  The consolidated result index gives exact pool sizes
for FSI, controlled cylinder, and foil, but not for cylinder.

\begin{table}[!htbp]
\centering
\caption{Additional Walrus full-pool endpoints.  Metrics are short/full.}
\label{tab:fullpool}
\small
\resizebox{\textwidth}{!}{%
\begin{tabular}{lrrlll}
\toprule
Dataset & Sweep maximum & Full-pool size & VRMSE & rel-$L_2$ & RMAE \\
\midrule
FSI & 32 & 36 & 0.0965/0.6446 & 0.0859/0.5462 & 0.0882/0.5042 \\
Cylinder & 64 & not recorded & 0.0214/0.3751 & 0.0162/0.2954 & 0.0120/0.2471 \\
Controlled cylinder & 64 & 76 & 0.0188/0.3711 & 0.0151/0.3016 & 0.0131/0.2602 \\
Foil & 64 & 79 & 0.1986/0.6819 & 0.1557/0.5633 & 0.0856/0.3708 \\
\bottomrule
\end{tabular}}
\end{table}

\FloatBarrier
\section{Compute and result provenance}
\label{app:compute}

Runs were executed on NCSA Delta (A100/H200, x86\_64) and DeltaAI (GH200, aarch64).  Every swept
cell retains a rollout archive containing predictions and ground truth together with its training
log.  Walrus and VICON retain per-trajectory prediction archives.  The common RealPDEBench
harvest and the DR/CondNS harvest implement the formulas in Section~\ref{sec:measurement} and use
the same checkpoint policy.  This retained provenance permits trajectory-level uncertainty and
physics-based diagnostics to be added without rerunning inference.

\section{Field visualizations}
\label{app:fields}

The tables above summarize each rollout with scalar error metrics.  This appendix shows the
underlying spatial fields so the qualitative behavior behind those numbers is visible.  For each
dataset we give two figures: a \emph{rollout comparison} (ground truth and every model across
prediction frames $t=10$--$20$ at the largest training budget) and a \emph{data-efficiency panel}
(one diagnostic frame, $t=20$, as the number of training trajectories $N$ grows).  Within a figure
all panels share one perceptually-uniform color scale taken from the ground-truth field
percentiles.  Each model is rolled out on its own native grid; the displayed trajectory, physical
channel, and absolute frame are matched to the ground truth by normalized cross-correlation, so a
given column is the same physical field in every row.  Walrus (full-split fine-tuned) and VICON
(in-context) are $N$-independent and therefore appear as fixed reference columns in the
data-efficiency panels.

\begin{figure}[p]
\centering
\figurepanelgraphic{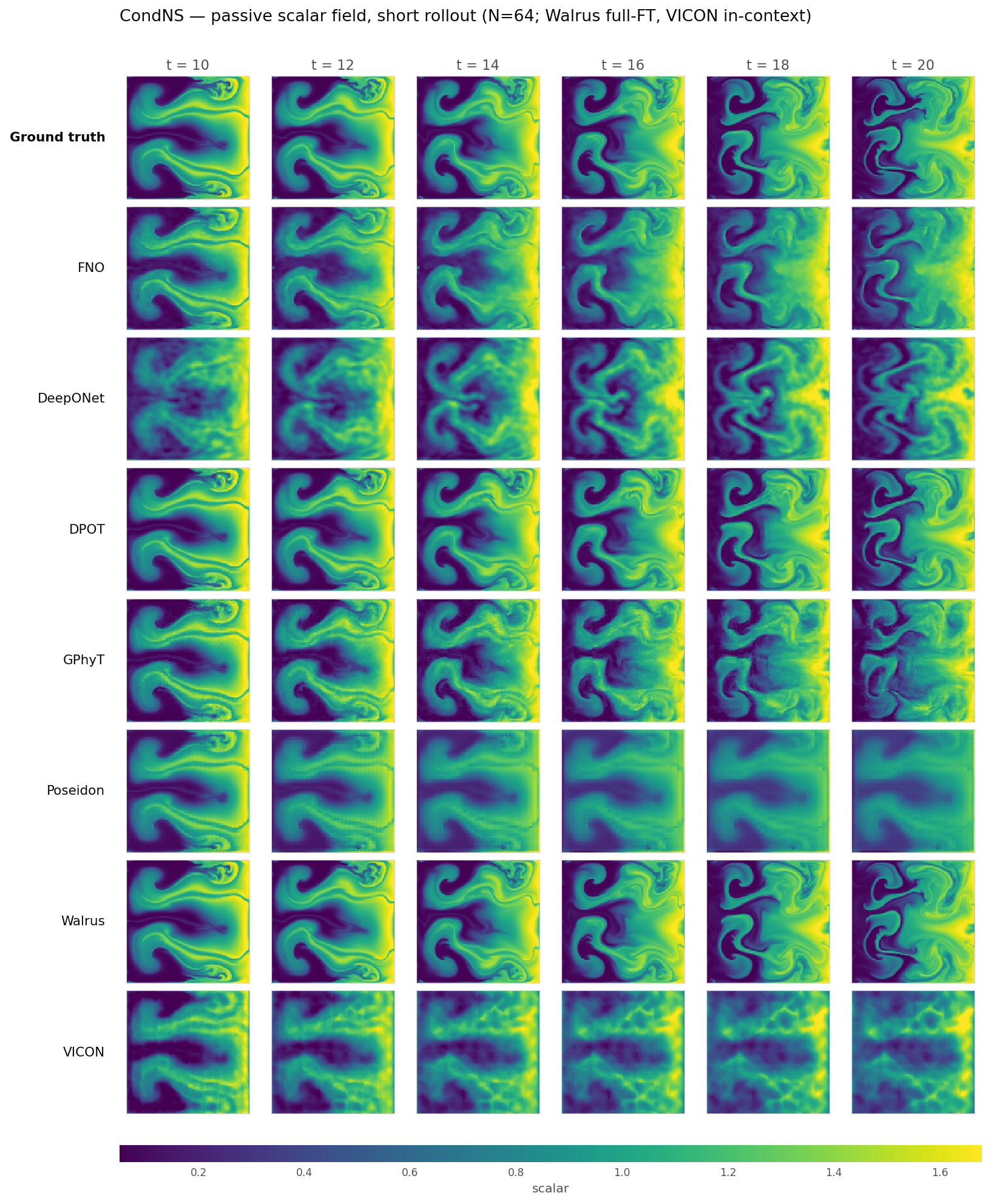}
\caption{\textbf{CondNS (passive scalar) --- rollout comparison.} Ground truth and all seven models
across frames $t=10$--$20$ at $N=64$.  Visual sharpness tracks the reported relative-$L_2$: Walrus
and DPOT retain the fine filaments, whereas Poseidon, DeepONet, and VICON diffuse or coarsen.}
\label{fig:field-condns-rollout}
\end{figure}

\begin{figure}[p]
\centering
\figurepanelgraphic{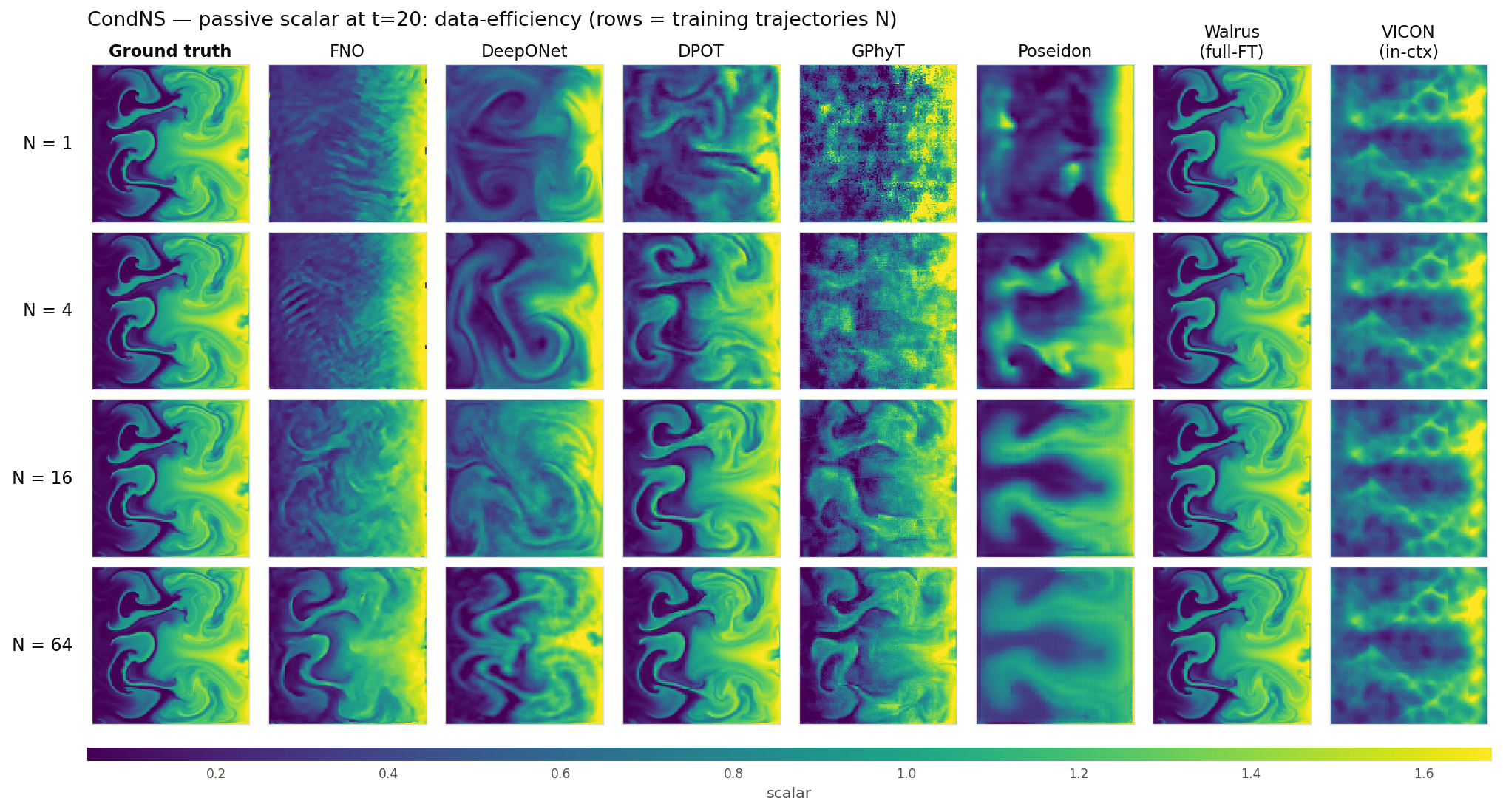}
\caption{\textbf{CondNS (passive scalar) --- data efficiency.} The five swept models at $t=20$ as
$N$ increases (rows), with the $N$-independent Walrus and VICON reference columns at right.}
\label{fig:field-condns-nsweep}
\end{figure}

\begin{figure}[p]
\centering
\figurepanelgraphic{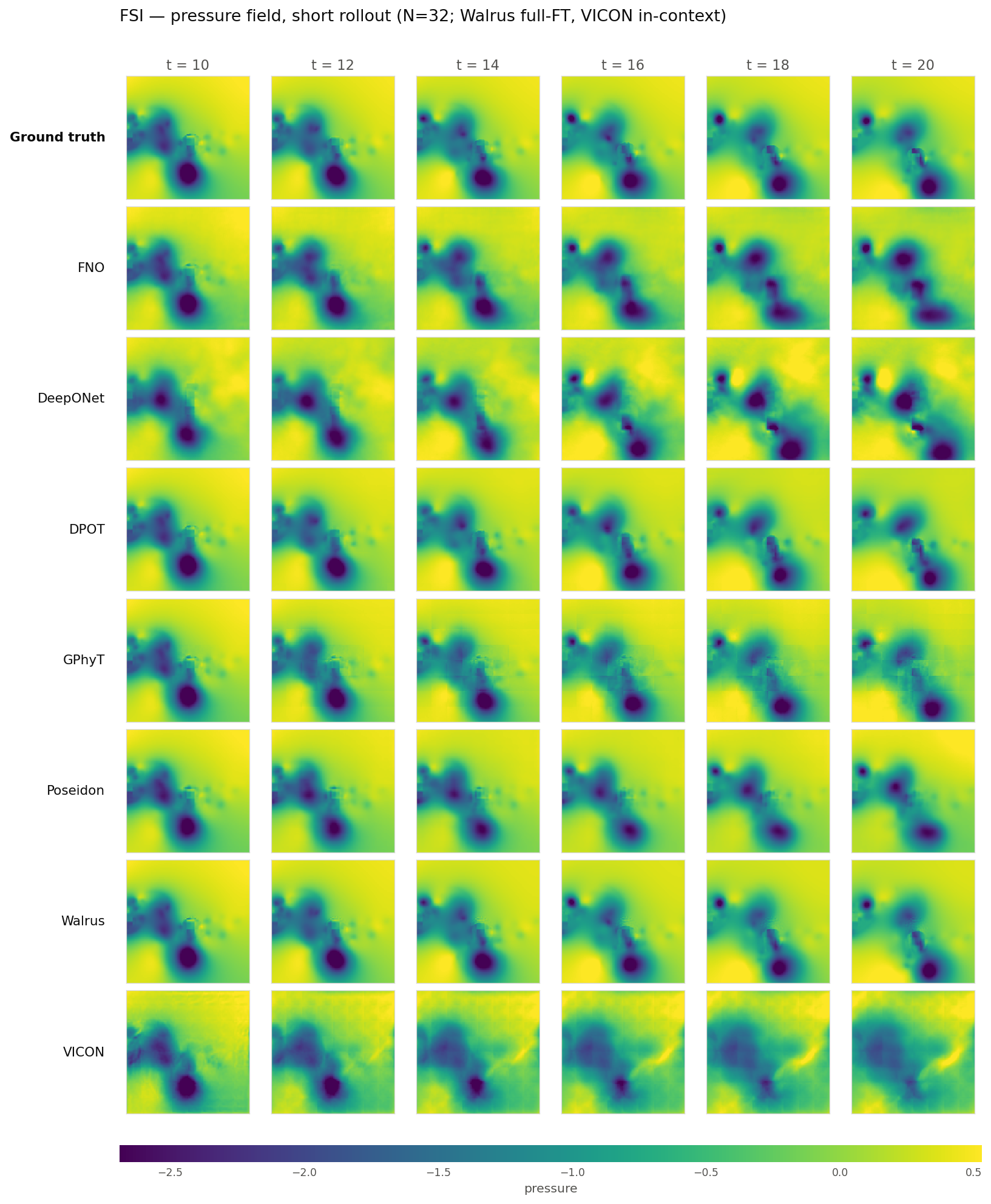}
\caption{\textbf{FSI (pressure) --- rollout comparison.} Ground truth and all seven models across
frames $t=10$--$20$ at $N=32$.  The moving low-pressure cores are tracked closely by Walrus, FNO,
DPOT, and Poseidon; VICON coarsens and loses structure by late frames.}
\label{fig:field-fsi-rollout}
\end{figure}

\begin{figure}[p]
\centering
\figurepanelgraphic{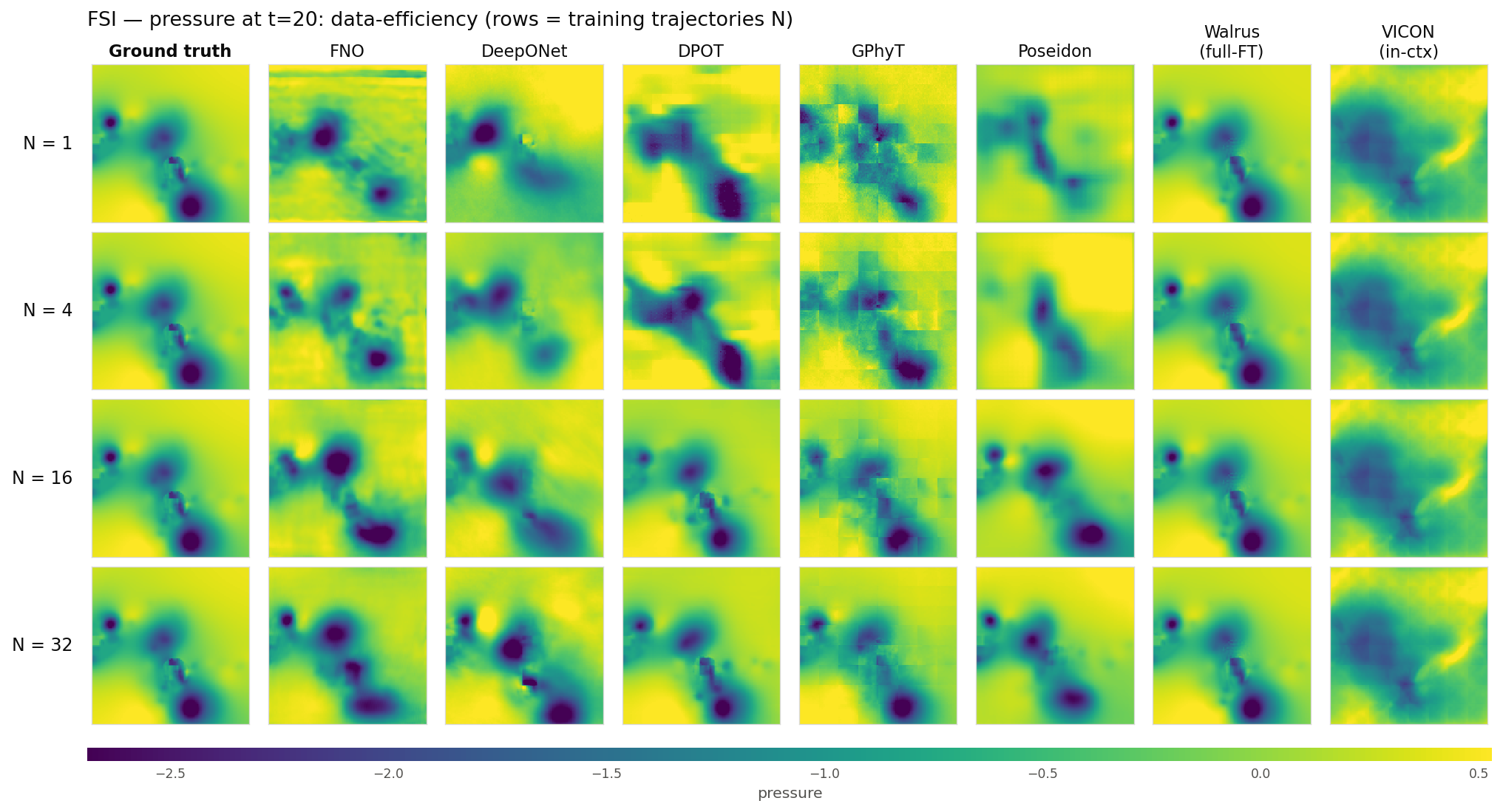}
\caption{\textbf{FSI (pressure) --- data efficiency.} Swept models at $t=20$ versus $N$, with Walrus
and VICON reference columns.  GPhyT is grainy at $N{=}1$ and sharpens with more trajectories.}
\label{fig:field-fsi-nsweep}
\end{figure}

\begin{figure}[p]
\centering
\figurepanelgraphic{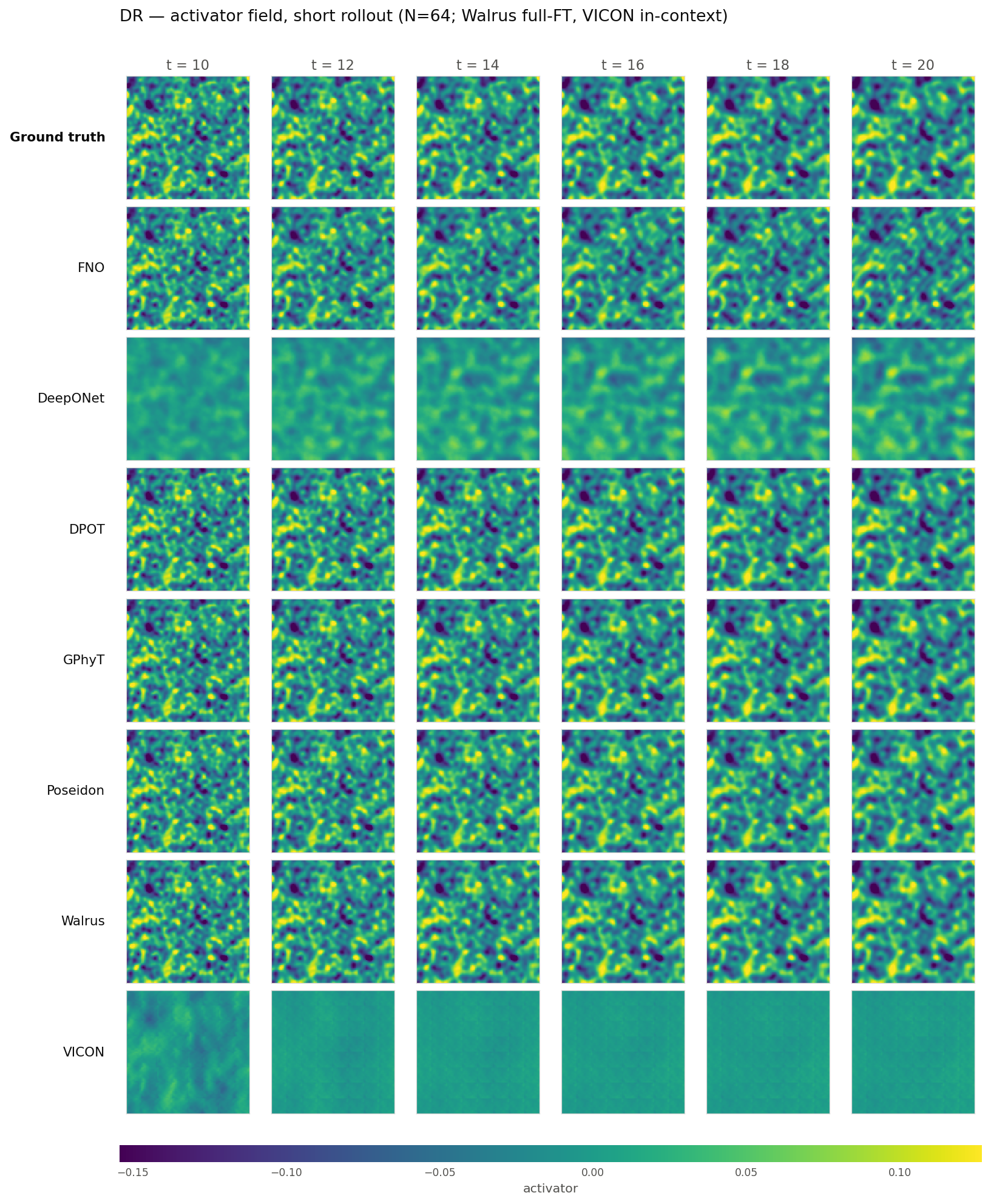}
\caption{\textbf{DR (activator) --- rollout comparison.} Ground truth and all seven models across
frames $t=10$--$20$ at $N=64$.  DPOT, GPhyT, Poseidon, and Walrus reproduce the Turing pattern;
DeepONet blurs it and VICON collapses to a near-constant field after the seeded frame.}
\label{fig:field-dr-rollout}
\end{figure}

\begin{figure}[p]
\centering
\figurepanelgraphic{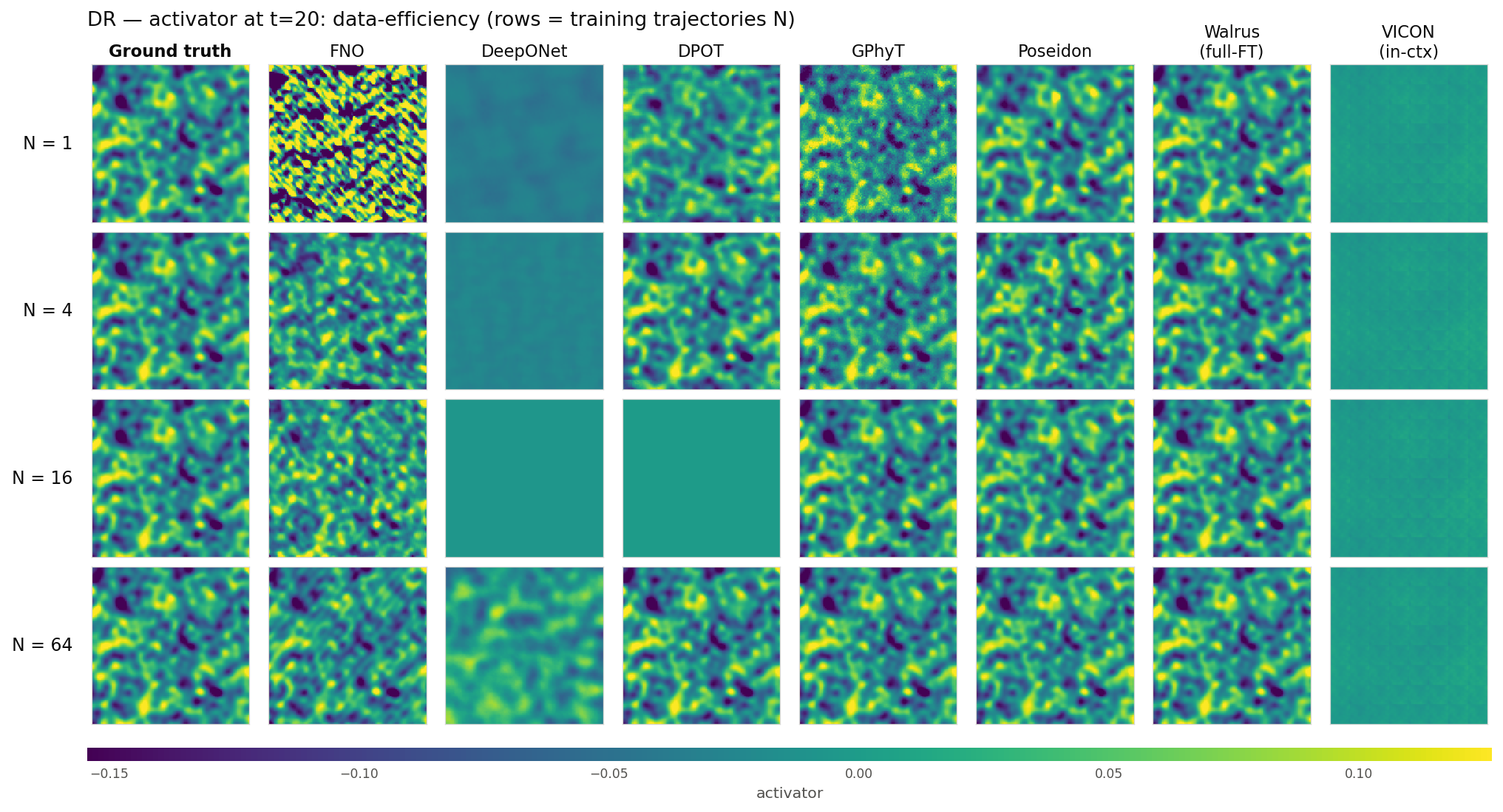}
\caption{\textbf{DR (activator) --- data efficiency.} Swept models at $t=20$ versus $N$, with Walrus
and VICON reference columns.  DeepONet is unstable (a flat or blobby field at several $N$); VICON is
$N$-independent and remains collapsed.}
\label{fig:field-dr-nsweep}
\end{figure}

\begin{figure}[p]
\centering
\figurepanelgraphic{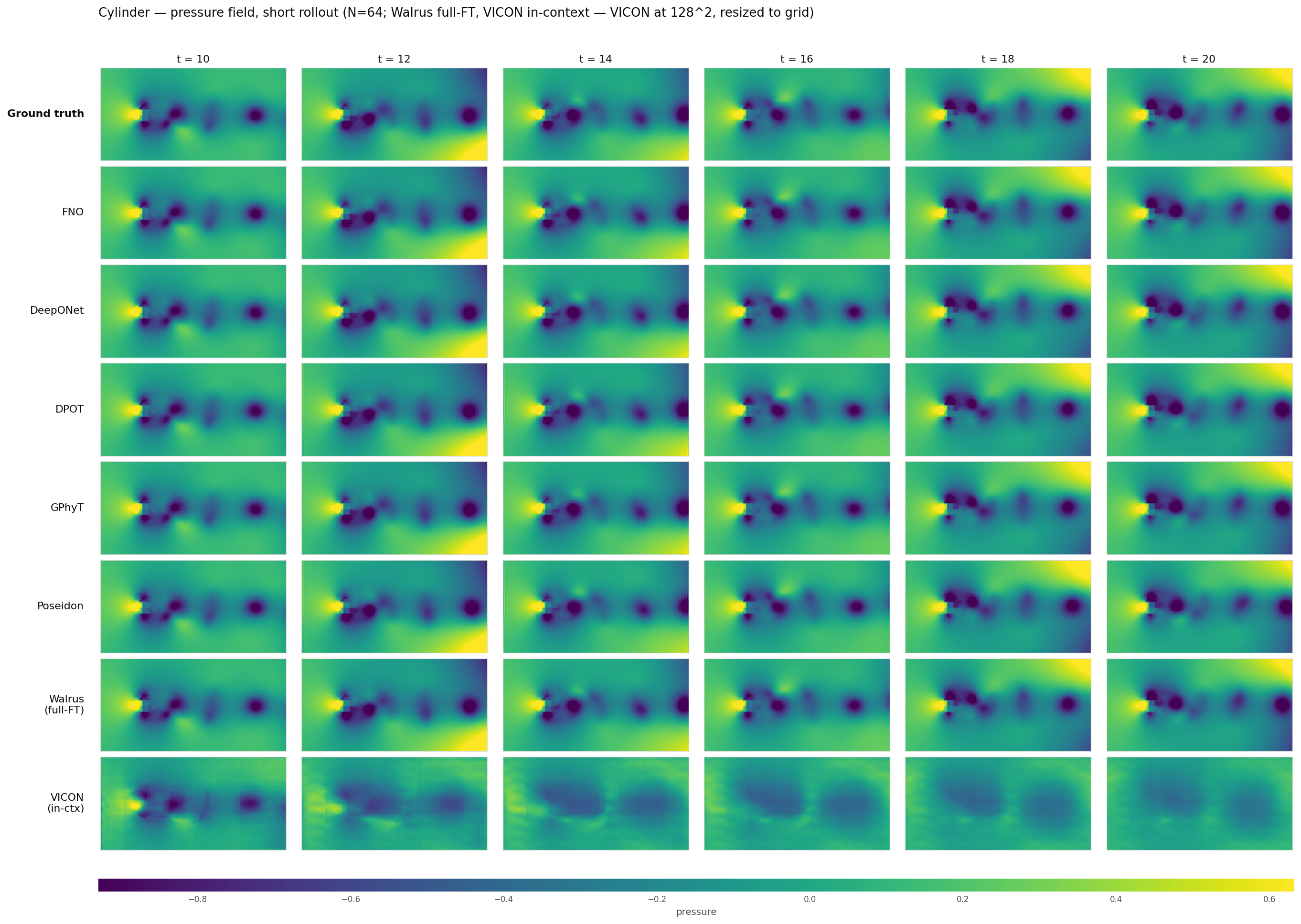}
\caption{\textbf{Cylinder (pressure) --- rollout comparison.} Ground truth and all seven models
across frames $t=10$--$20$ at $N=64$ on the native $128\times256$ grid.  The von K\'arm\'an vortex
street is tracked by most trained models; VICON, which operates at $128^2$ and is resized to the
grid for display, is diffuse.}
\label{fig:field-cyl-rollout}
\end{figure}

\begin{figure}[p]
\centering
\figurepanelgraphic{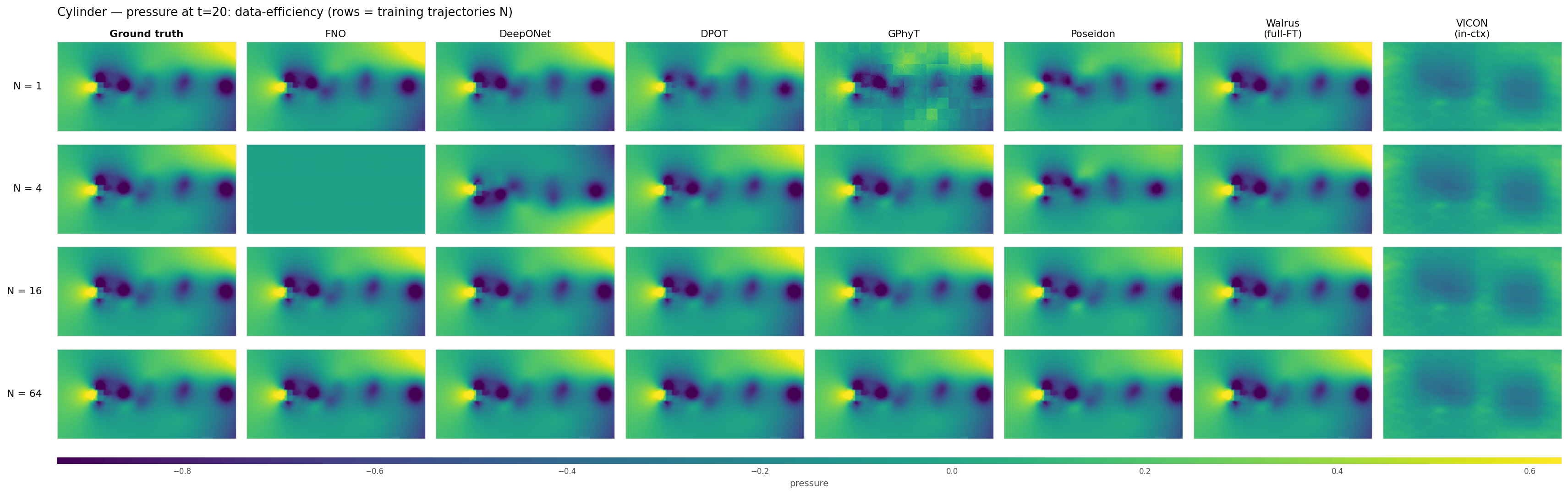}
\caption{\textbf{Cylinder (pressure) --- data efficiency.} Swept models at $t=20$ versus $N$, with
Walrus and VICON reference columns.}
\label{fig:field-cyl-nsweep}
\end{figure}

\begin{figure}[p]
\centering
\figurepanelgraphic{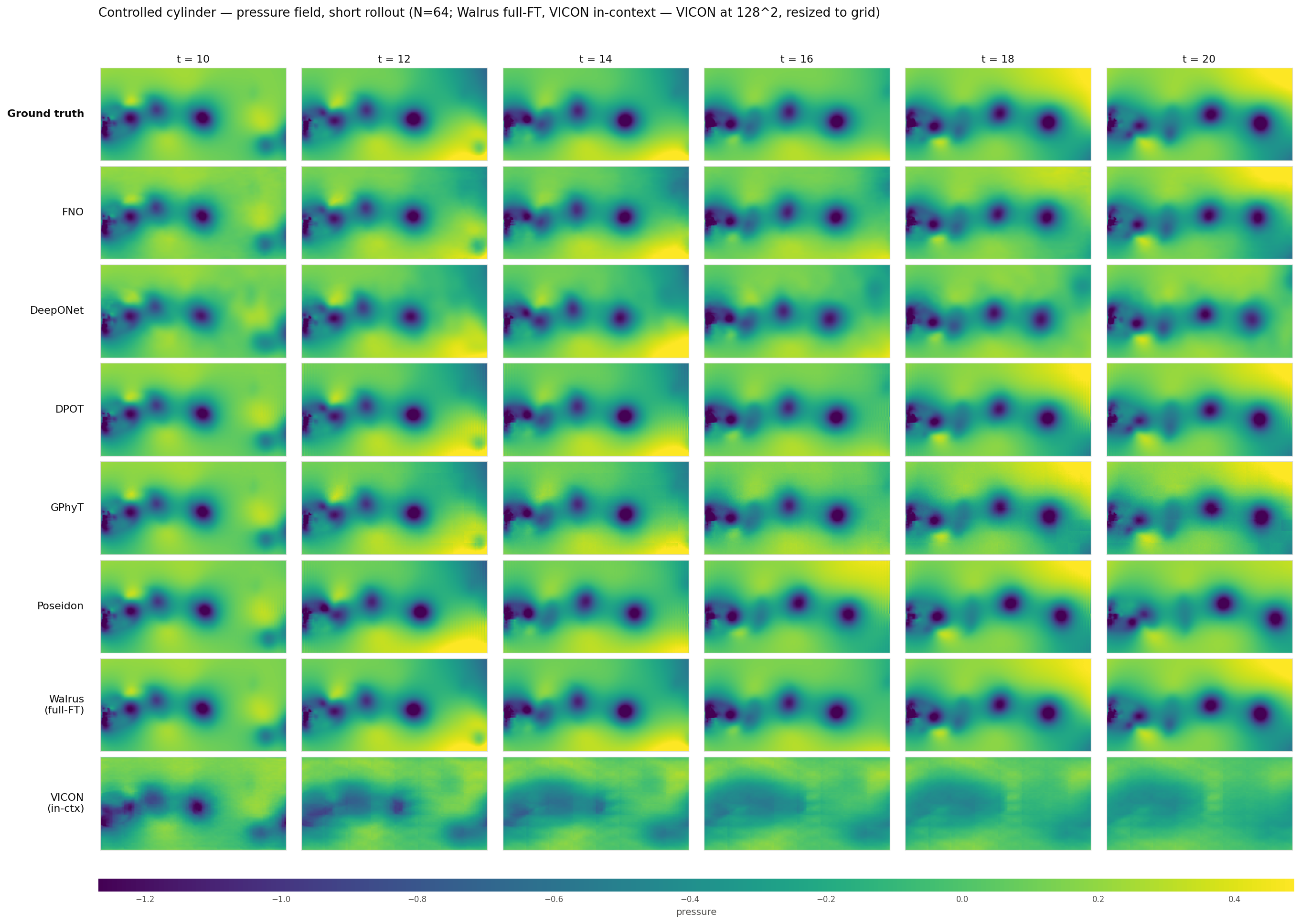}
\caption{\textbf{Controlled cylinder (pressure) --- rollout comparison.} Ground truth and all seven
models across frames $t=10$--$20$ at $N=64$ ($128\times256$; VICON resized from $128^2$).}
\label{fig:field-ctrlcyl-rollout}
\end{figure}

\begin{figure}[p]
\centering
\figurepanelgraphic{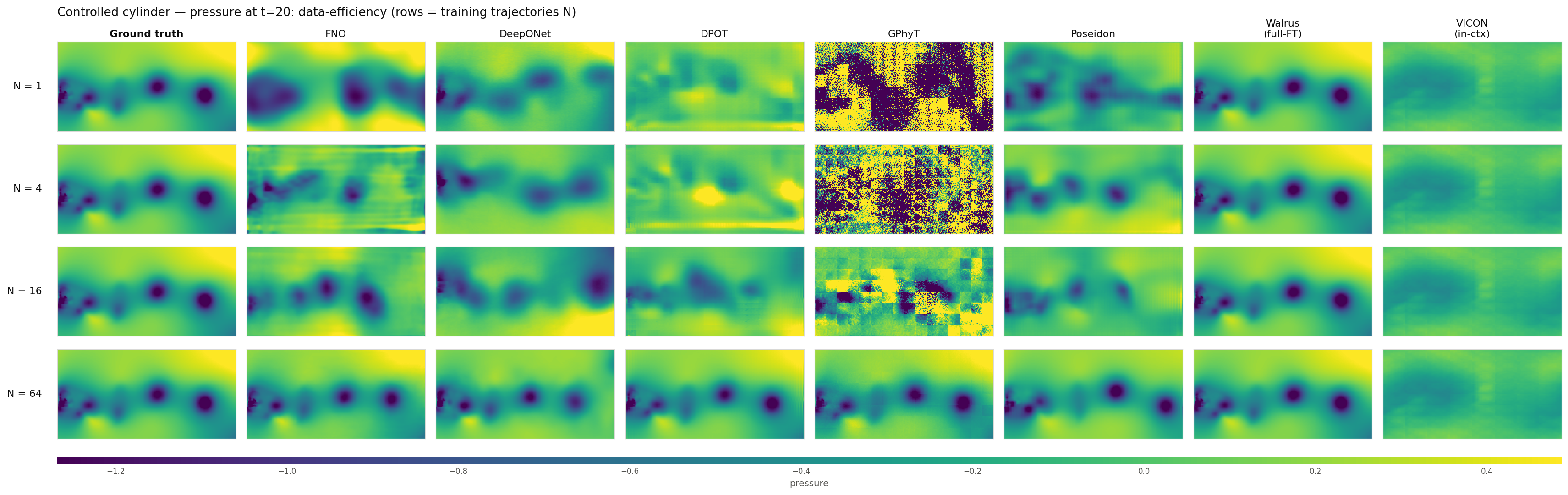}
\caption{\textbf{Controlled cylinder (pressure) --- data efficiency.} Swept models at $t=20$ versus
$N$, with Walrus and VICON reference columns.}
\label{fig:field-ctrlcyl-nsweep}
\end{figure}

\begin{figure}[p]
\centering
\figurepanelgraphic{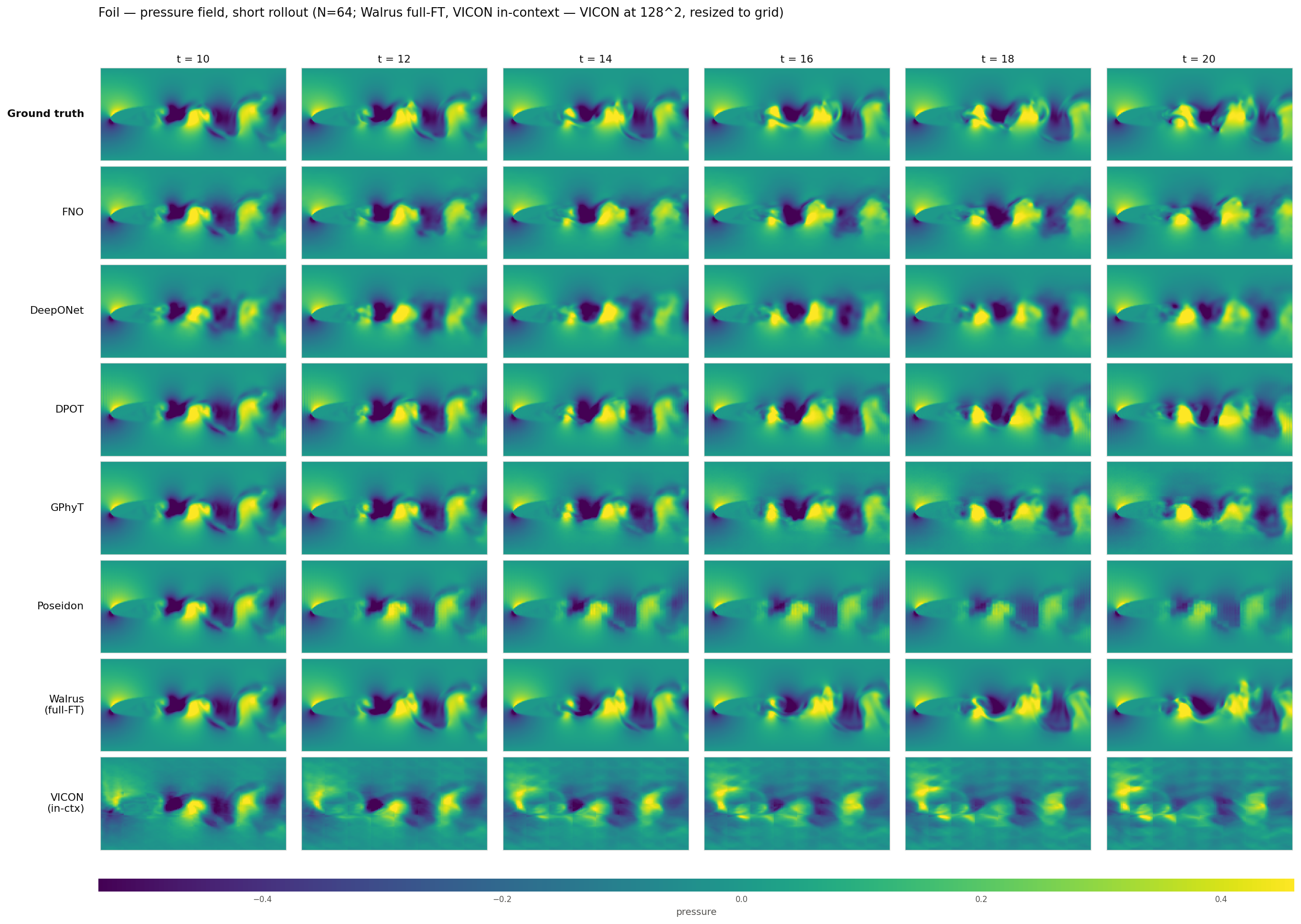}
\caption{\textbf{Foil (pressure) --- rollout comparison.} Ground truth and all seven models across
frames $t=10$--$20$ at $N=64$ ($128\times256$; VICON resized from $128^2$).}
\label{fig:field-foil-rollout}
\end{figure}

\begin{figure}[p]
\centering
\figurepanelgraphic{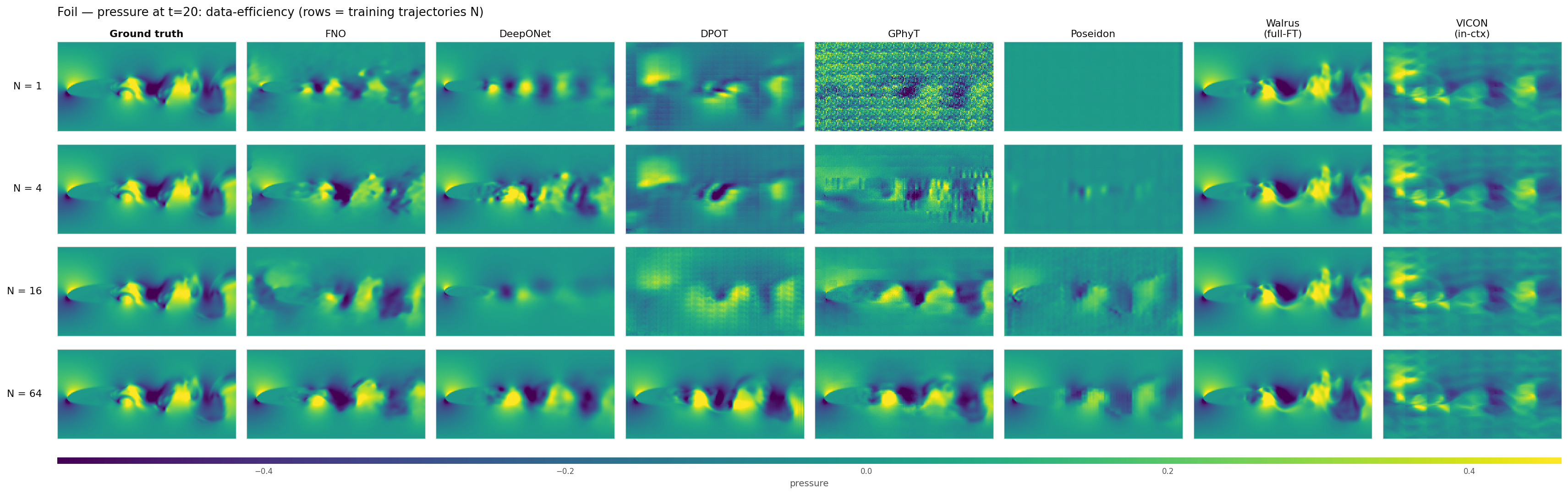}
\caption{\textbf{Foil (pressure) --- data efficiency.} Swept models at $t=20$ versus $N$, with
Walrus and VICON reference columns.}
\label{fig:field-foil-nsweep}
\end{figure}

\begin{figure}[p]
\centering
\figurepanelgraphic{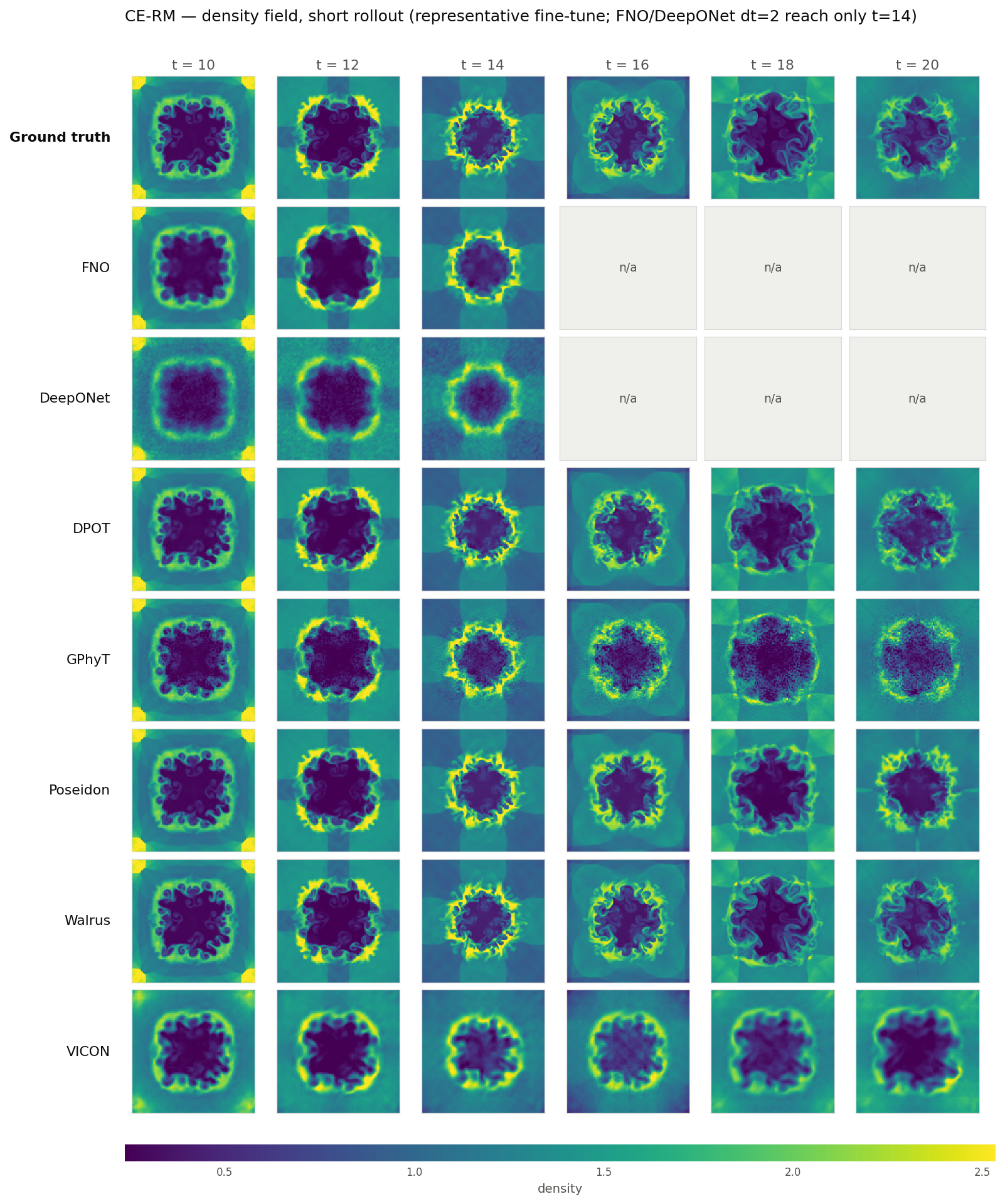}
\caption{\textbf{CE-RM (density) --- rollout comparison.} Richtmyer--Meshkov interface for all seven
models across frames $t=10$--$20$.  FNO and DeepONet advance at $\Delta t=2$ from an early context
window and reach only $t=14$, so their later cells are marked \texttt{n/a}.  CE-RM has no
data-efficiency panel because per-$N$ rollout archives were not retained for every swept model.}
\label{fig:field-cerm-rollout}
\end{figure}

\section{Per-frame rollout error curves}
\label{app:errorcurves}

Where the field figures show \emph{how} each rollout looks, these curves show \emph{when} error
accumulates.  For every dataset we plot the three metrics (VRMSE, relative-$L_2$, RMAE) against the
absolute prediction frame, one line per model, at the largest training budget (Walrus full-split
fine-tuned, VICON in-context).  The per-frame value is the median over test trajectories of the
channel-averaged error at that frame; averaging a curve over the short window $t=10$--$20$
reproduces the corresponding scalar entry in the tables above to within one to two percent.  The
vertical axis is logarithmic.  On CE-RM the models do not share a frame grid: FNO and DeepONet
advance at $\Delta t=2$ from an early context window and terminate at $t=14$, Poseidon steps at
$\Delta t=2$ to $t=20$, and the remaining models step at $\Delta t=1$; frames are assigned by
matching each model's ground truth to the reference by normalized cross-correlation.

\begin{figure}[p]
\centering
\figurepanelgraphic{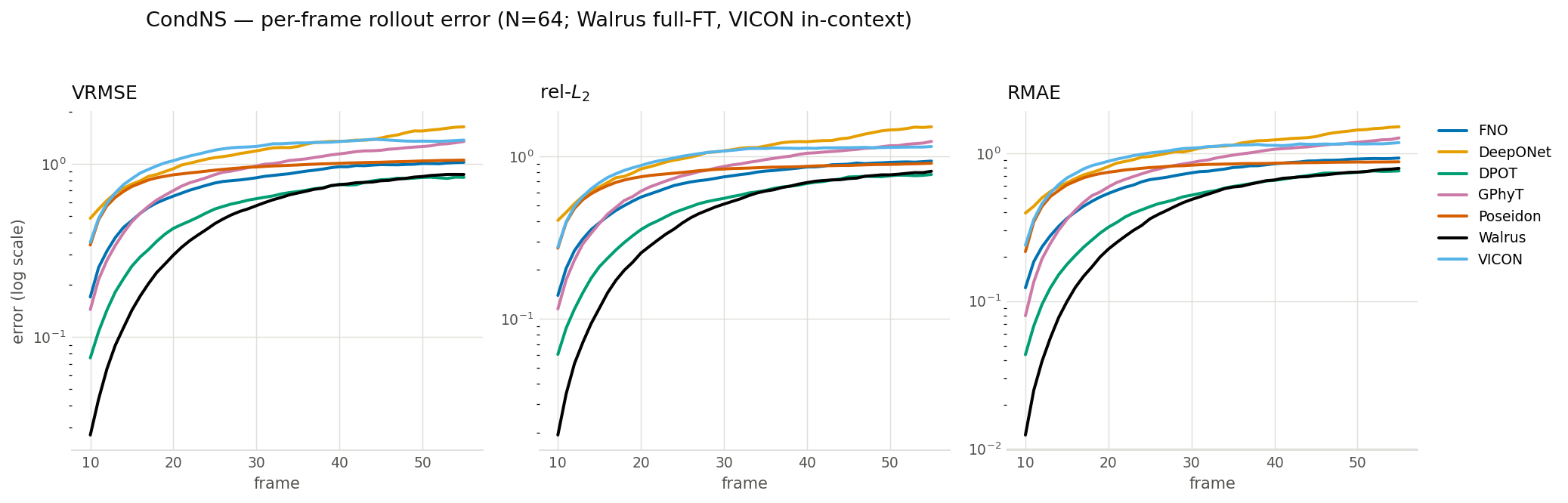}
\caption{\textbf{CondNS --- per-frame rollout error.} All three metrics versus prediction frame at
$N=64$.  Every model amplifies error monotonically and saturates by mid-rollout; Walrus stays lowest
throughout and DeepONet and VICON saturate near unity.}
\label{fig:err-condns}
\end{figure}

\begin{figure}[p]
\centering
\figurepanelgraphic{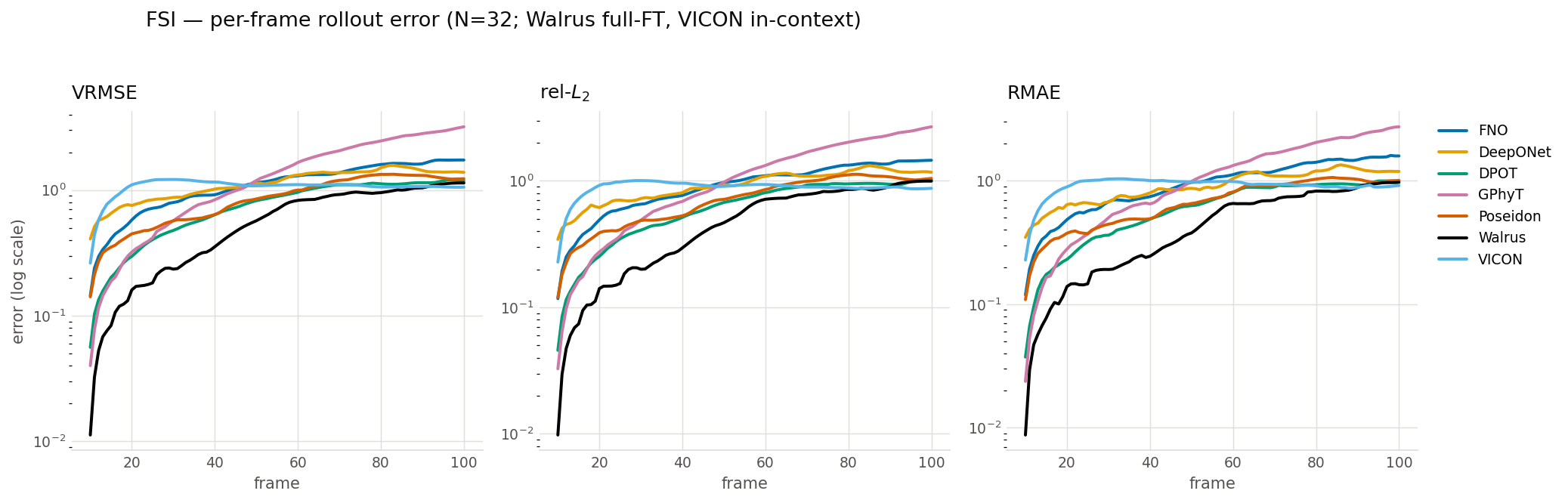}
\caption{\textbf{FSI --- per-frame rollout error.} Metrics versus frame at $N=32$.  DPOT and GPhyT
track Walrus most closely; VICON is nearly flat and high, reflecting its frozen in-context inference.}
\label{fig:err-fsi}
\end{figure}

\begin{figure}[p]
\centering
\figurepanelgraphic{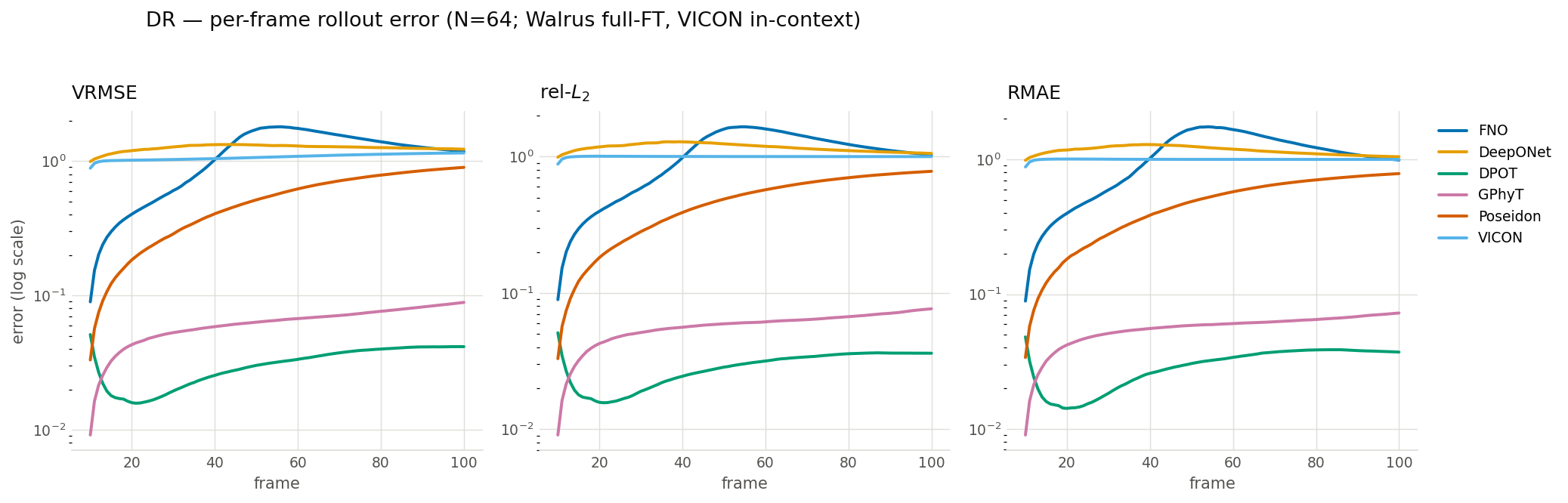}
\caption{\textbf{DR --- per-frame rollout error.} Metrics versus frame at $N=64$.  DPOT, GPhyT, and
Poseidon remain low across the horizon, whereas DeepONet and VICON sit near unity (no better than
predicting the mean field) from the first predicted frame.}
\label{fig:err-dr}
\end{figure}

\begin{figure}[p]
\centering
\figurepanelgraphic{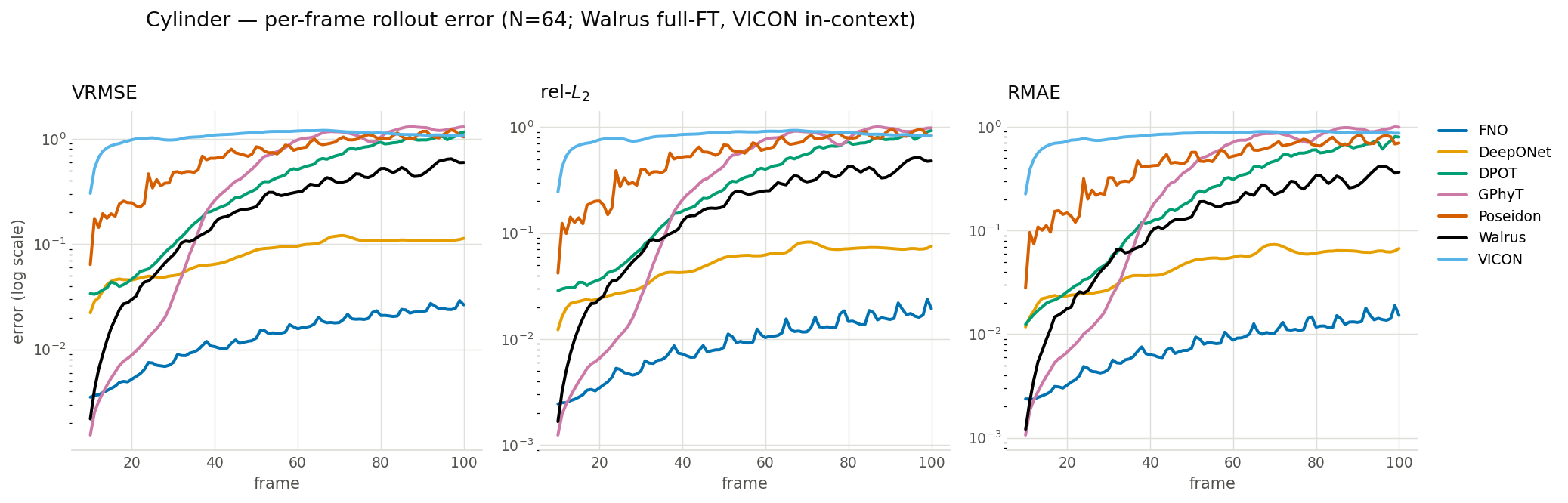}
\caption{\textbf{Cylinder --- per-frame rollout error.} Metrics versus frame at $N=64$.  The periodic
wake is the easiest target: most trained models stay well below $0.1$, with VICON the clear outlier.}
\label{fig:err-cyl}
\end{figure}

\begin{figure}[p]
\centering
\figurepanelgraphic{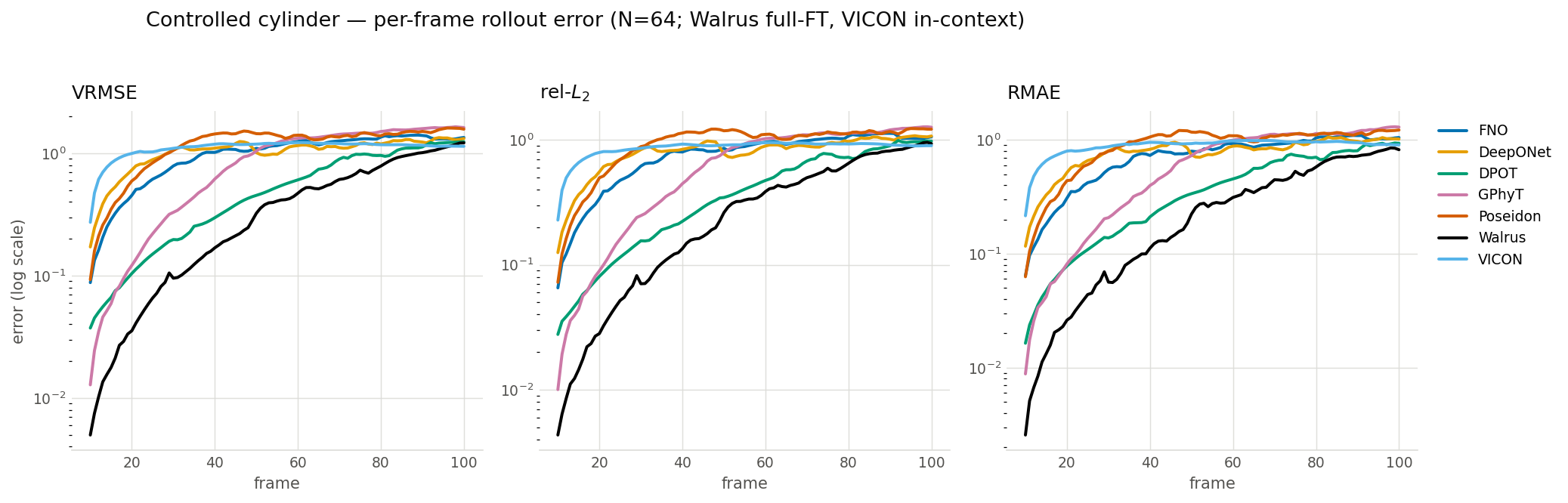}
\caption{\textbf{Controlled cylinder --- per-frame rollout error.} Metrics versus frame at $N=64$.
Walrus, DPOT, and GPhyT remain low; Poseidon and DeepONet amplify faster.}
\label{fig:err-ctrlcyl}
\end{figure}

\begin{figure}[p]
\centering
\figurepanelgraphic{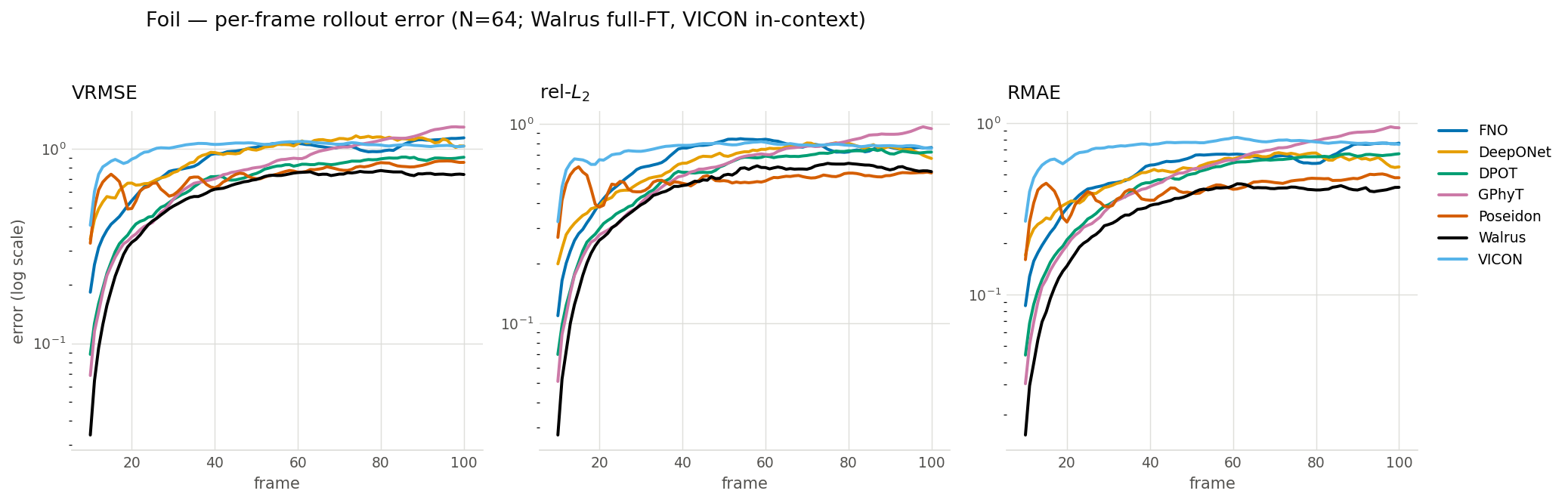}
\caption{\textbf{Foil --- per-frame rollout error.} Metrics versus frame at $N=64$.}
\label{fig:err-foil}
\end{figure}

\begin{figure}[p]
\centering
\figurepanelgraphic{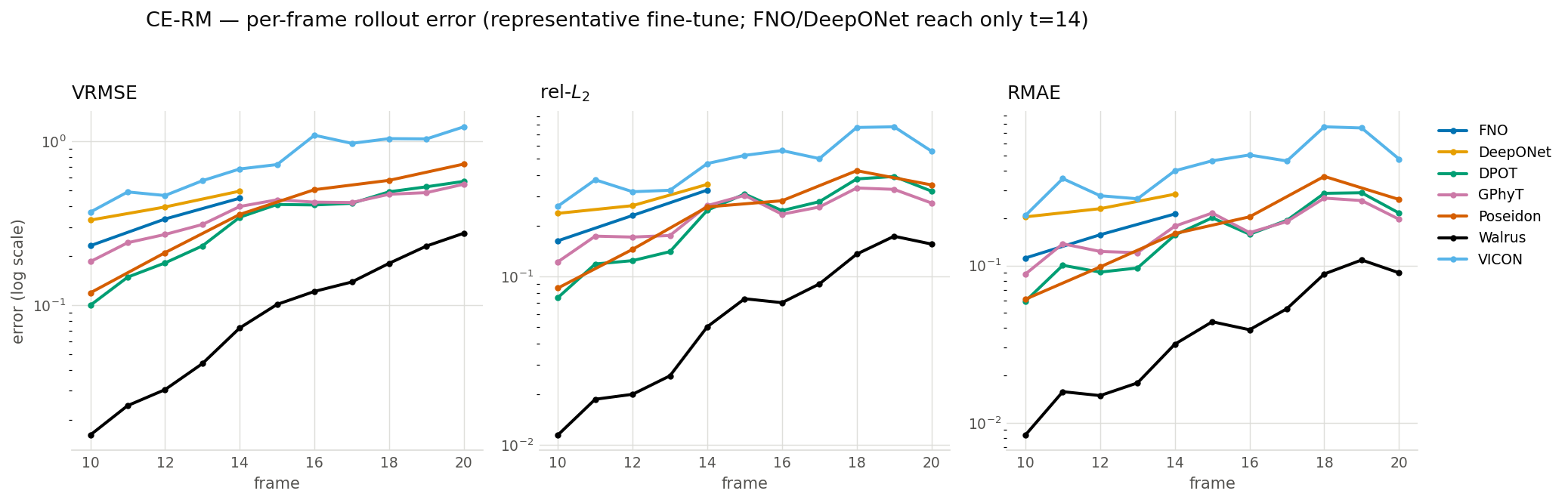}
\caption{\textbf{CE-RM --- per-frame rollout error.} Metrics versus frame for the representative
fine-tune.  Markers denote the frames each model actually predicts: FNO and DeepONet ($\Delta t=2$
from an early window) terminate at $t=14$, Poseidon reaches $t=20$ at $\Delta t=2$, and the
remaining models step at $\Delta t=1$.  Walrus again stays lowest throughout.}
\label{fig:err-cerm}
\end{figure}

\end{document}